\documentclass{article}

\usepackage{microtype}
\usepackage{graphicx}
\usepackage{subcaption}
\usepackage{booktabs} 

\usepackage{hyperref}

\usepackage[preprint]{icml2026}

\usepackage{amsmath}
\usepackage{amssymb}
\usepackage{mathtools}
\usepackage{amsthm}

\usepackage[capitalize,noabbrev]{cleveref}

\usepackage{float}          
\floatstyle{plain}          

\usepackage[section]{placeins}

\usepackage{mathtools}
\usepackage{booktabs}
\usepackage{colortbl}
\usepackage{array}
\usepackage{arydshln} 
\usepackage{tikz}
\usetikzlibrary{arrows.meta, positioning}
\usepackage{subcaption}

\usepackage{booktabs}
\usepackage{array}
\usepackage[table]{xcolor}

\usepackage{amsfonts}
\usepackage{amssymb}

\usepackage{bm} 
\usepackage[table]{xcolor}
\usepackage{booktabs}
\usepackage{multirow}
\usepackage{array}
\newcommand{\ContextSet}{C_k}
\newcommand{\Params}{\boldsymbol{\theta}} 
\newcommand{\ParamsIc}{\boldsymbol{\theta}_c} 
\newcommand{\LogitsM}{\mathbf{m}} 
\newcommand{\LogitsMc}{m_c} 
\newcommand{\TrueLogitsL}{L^{\star}_c} 

\newcommand{\SizeSetI}{I} 

\newcommand{\RegLambda}{\lambda_{\text{inv}}}

\newcommand{\LearnedParams}{\hat{\Params}} 

\newtheorem{definition}{Definition}
\newcommand{\MethodName}{SC}

\usepackage{hyperref}       
\usepackage{url}            
\usepackage{booktabs}       
\usepackage{amsfonts}       
\usepackage{nicefrac}       
\usepackage{microtype}      
\usepackage{xcolor}      

\usepackage{etoolbox}   

\newtheorem{assumption}{Assumption}

\theoremstyle{plain}

\theoremstyle{definition}
\theoremstyle{remark}

\usepackage[disable,textsize=tiny]{todonotes}

\icmltitlerunning{Boosting In-Context Learning in LLMs Through the Lens of Classical Supervised Learning}

\begin{document}

\twocolumn[
  \icmltitle{Boosting In-Context Learning in LLMs Through the Lens of Classical Supervised Learning}



  \icmlsetsymbol{equal}{*}

  \begin{icmlauthorlist}
  \icmlauthor{Korel Gundem}{equal,gwu}
  \icmlauthor{Juncheng Dong}{equal,duke}
  \icmlauthor{Dennis Zhang}{wustl}
  \icmlauthor{Vahid Tarokh}{duke}
  \icmlauthor{Zhengling Qi}{gwu}
\end{icmlauthorlist}

\icmlaffiliation{gwu}{Department of Decision Sciences, The George Washington University, Washington, DC, USA}
\icmlaffiliation{duke}{Duke University, Durham, NC, USA}
\icmlaffiliation{wustl}{Washington University in St. Louis, St. Louis, MO, USA}

  \icmlcorrespondingauthor{Korel Gundem}{korelgundem@gwu.edu}

  \icmlkeywords{Machine Learning, ICML}

  \vskip 0.3in
]



\printAffiliationsAndNotice{}  

\begin{abstract}
  In-Context Learning (ICL) allows Large Language Models (LLMs) to adapt to new tasks with just a few examples, but their predictions often suffer from systematic biases, leading to unstable performance in classification. While calibration techniques are proposed to mitigate these biases, we show that, in the logit space, many of these methods are equivalent to merely shifting the LLM's decision boundary without having the ability to alter its orientation. This proves inadequate when biases cause the LLM to be severely misaligned. To address these limitations and provide a unifying framework, we propose Supervised Calibration (SC), a loss-minimization-based framework, which learns an optimal, per-class affine transformation of LLM's predictive probabilities in the logit space without requiring external data beyond the context. By using a more expressive functional class, SC not only subsumes many existing calibration methods in ICL as special cases but also enables the ability of altering and even completely reversing the orientation of the LLM's decision boundary. Furthermore, SC's loss-based nature facilitates the seamless integration of two purpose-built regularization techniques, context-invariance and directional trust-region regularizers. The former is designed to tackle the instability issue in ICL, while the latter is to control the degree of calibration. Finally, SC delivers state-of-the-art performance over calibration baselines in the 4-shot, 8-shot, and 16-shot settings across all nine datasets for Mistral-7B-Instruct-v0.3, Llama-2-7B-chat, and Qwen2-7B-Instruct.
\end{abstract}

\section{Introduction}
State‑of‑the‑art LLMs exhibit a striking \textit{in‑context learning} (ICL) capability: with only a handful of input–label exemplars, they generalize to unseen queries almost as if they had been fine‑tuned, thus functioning as highly sample‑efficient few‑shot learners \citep{brown2020languagemodelsfewshotlearners,liu2021pretrainpromptpredictsystematic}.  
However, a growing body of evidence shows that ICL performance can be  brittle with respect to seemingly innocent design choices such as template wording \citep{min2022rethinkingroledemonstrationsmakes,voronov2024mind} and the particular demonstrations given \citep{qin2024context}.  
These biases and sensitivities of ICL pose a practical barrier to developing applications that are both adaptable and robust. Motivated by this, extensive research has been conducted to develop calibration approaches to address such a challenge for classification problems in ICL. The majority of calibration methods fall under label-marginal-based calibration (LM). These methods first estimate the LLM's probability for each label given the context alone via various approaches. 
They then discount the predictive probabilities of the LLM for the labels that are over-represented and boost those that are under-represented. See detailed discussion in the later sections.

\begin{figure*}[t]
    \centering
    \includegraphics[width=0.75\textwidth]{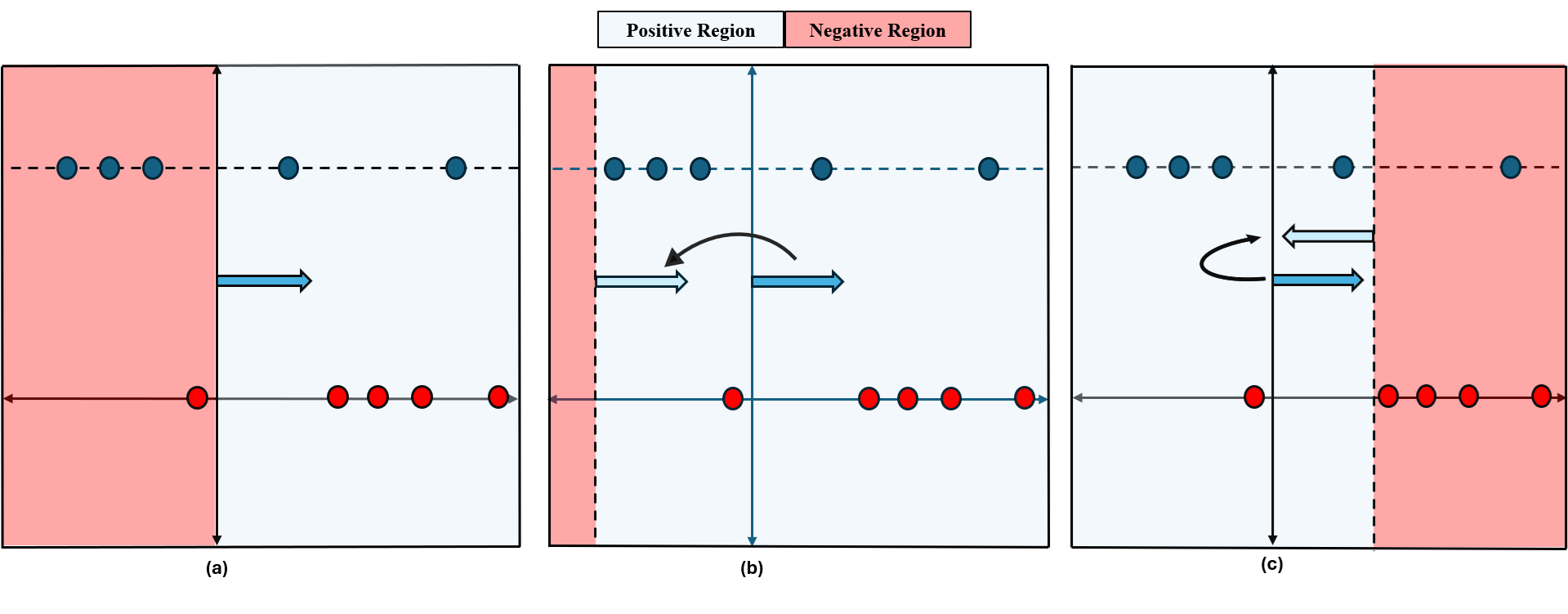}
    \caption{
        Comparison of ICL prediction strategies, where the x-axis represents the LLM’s raw logits (log-odds). 
        \textbf{(a)} \textit{Base LLM (accuracy: 30\%)}: The model predicts class 1 when $\text{logit} > 0$.
        \textbf{(b)} \textit{Label Marginal Calibration (accuracy: 50\%)}: These methods only shift the decision boundary,  limiting correction when base LLM is systematically wrong.
        \textbf{(c)} \textit{Supervised Calibration (accuracy: 80\%)}: SC can shift and flip the decision boundary of the base LLM, resulting in a significant improvement.}
    \label{fig:calibration_comparison}
\end{figure*}
Despite the empirical success of these methods, their ability of correcting the predictive probabilities of the LLM via its internal estimated prior is limited. Specifically, we show in Section~\ref{sec:theory_combined} that the underlying idea of these methods is equivalent to optimally shifting the decision threshold of the base LLM. Hence, they are inherently incapable of altering or reversing the orientation of the decision boundary. This becomes problematic when the base LLM performs poorly. To further illustrate this limitation, consider a binary classification problem in Figure \ref{fig:calibration_comparison}~(a), where the base LLM only achieves $30\%$ accuracy. Since LM methods can only shift the decision threshold, their maximum improvement over the base LLM is capped, only achieving the level of random guessing as seen in Figure \ref{fig:calibration_comparison}~(b). One may expect that such an issue becomes more common and severe in the multiclass classification, where distinguishing among a larger number of labels is inherently more difficult. For instance, on the \textsc{SST-5} dataset, the average accuracy across three representative LLMs is only 22\%, highlighting the severity of this challenge. This limitation motivates the need for a more principled calibration framework that is capable of correcting severely misaligned LLM predictions when necessary (e.g., by reversing the decision direction), and that subsumes existing methods as special cases while remaining both theoretically grounded and practically robust.


To achieve this goal,  we introduce Supervised Calibration (\MethodName{}), which is motivated by conceptualizing existing approaches as learning a calibrated classifier: they take a LLM's logits as input features and subsequently optimize a bias term to shift these logits. However, this shift only corresponds to moving the LLM's decision boundary to maximize  the predictive accuracy illustrated in Figure \ref{fig:calibration_comparison}~(b). Therefore, to enable more comprehensive adjustments, specifically, the ability to alter or reverse the orientation of the LLM's decision boundary, the proposed \MethodName{} leverages the paradigm of loss-function-based classification and optimizes both the bias and the scaling factor jointly. Our approach begins by generating a surrogate dataset, removing the need for an external dataset beyond the given context. From this surrogate data, we extract features in the form of logits derived from the base LLM’s output probabilities. Then we employ these features, paired with their corresponding true labels to train a standard classifier, which learns not only an optimal bias term but also an optimal rescaling factor. Critically, the concurrent optimization of this rescaling factor empowers our approach to reverse the LLM's decision boundary  when advantageous (as illustrated in Figure \ref{fig:calibration_comparison}~(c)). 
Moreover, the loss-minimization framework underpinning \MethodName{} inherently supports the integration of regularization techniques designed for addressing the common problems in ICL and calibration. In this context, we propose  a novel context-invariance regularizer for addressing the instability issue in ICL and a directional trust-region regularizer for controlling the degree of calibration. From a statistical viewpoint, these characteristics allow \MethodName{} to pursue a balance with respect to the bias–variance trade-off. While SC's flexibility targets a reduction in approximation error over LM methods, its regularization components actively constrain variance which is an essential consideration within the data-scarce ICL paradigm. Collectively, \MethodName{} delivers an adaptable, stable, and theoretically grounded framework that improves LLMs' classification quality in few-shot settings, potentially improving robustness across prompts and datasets as a result. Experimental results demonstrate that \MethodName{} consistently outperforms existing calibration methods across a broad range of tasks, significantly enhancing the predictive performance of three distinct LLMs evaluated on nine inference datasets. For example, the performance of \MethodName{} is striking on the SST-5 dataset  with the Qwen model (8-shot setting), where it significantly outperforms baseline methods with accuracy from ~25\% (baselines) to 44\%. This notable boost is directly attributable to its learned negative scaling factor which re-orients the base LLM decision boundary in this multiclass classification task. See Figure \ref{fig:sst5} for more details. 

Our main contributions are summarized as follows: Firstly, we propose Supervised Calibration, which adopts a loss minimization framework from classical supervised learning and calibrates ICL via learning optimal bias and scaling factors, enabling not only shifting but also altering the orientation of the base LLM decision boundary; Secondly, we integrate the context-invariance and directional trust-region regularizations in SC, enhancing the  stability of ICL and controlling the degree of the calibration respectively; Thirdly, we provide a theoretical intuition behind SC and its generalization over the LM methods; Lastly, we conduct extensive empirical studies to demonstrate the state-of-the-art performance of SC over several existing baselines.\footnote{The code for reproducibility: \url{https://github.com/gundemkorel/ICL}}
\section{Related Work}
\textbf{Diagnosing biases and calibration via Label Marginal}.
A seminal study by \citet{zhao2021calibrate} identified primary in-context learning (ICL) biases—including majority-label, recency, and common-token bias—and introduced \textbf{Contextual Calibration (CC)}, which adjusts probabilities by normalizing against content-free prompts. Subsequently, observing that competition for probability mass degrades performance, \citet{holtzman2022surfaceformcompetitionhighest} proposed \textbf{DCPMI} to recalibrate logits. Recent work has uncovered further ICL instabilities, such as feature and positional biases, with each diagnosis often paired with a lightweight calibration strategy \citep{si2023measuring, wang2023largelanguagemodelsfair, pezeshkpour2023sensitivity}. For instance, \textbf{Domain-Context Calibration (DC)} corrects predictions by averaging over random in-domain strings \citep{fei2023mitigating}, while the more recent \textbf{Batch Calibration (BC)} uses unlabeled mini-batches to adjust each prediction \citep{zhou2023batch}. Although these methods show empirical improvements, they can fail when the base LLM is substantially misaligned with the downstream task, as they cannot alter the model's decision direction. This limitation motivates the exploration of calibration frameworks with greater flexibility.

\textbf{Calibration via centroids.} A parallel line of work mitigates in-context biases by replacing the standard decision rule with centroid-based classification. \citet{han2022prototypical} proposed \textbf{Prototypical Calibration}, which models output probability vectors using Gaussian mixtures and assigns labels based on cluster likelihood, improving robustness to prompt variation and class imbalance. Similarly, \citet{cho2024token} introduced \textbf{Hidden Calibration}, which operates in the model’s latent space by computing class centroids over hidden states and classifying based on proximity. Although these methods show empirical performance gains, they rely on additional data beyond the in-context examples, which may not always be available or compatible with the ICL setting.

\section{Supervised Calibration}
\subsection{Background}
Consider an $n$‑class classification task with label verbaliser set $\mathcal{Y} = \{y_0, \dots, y_{n-1}\}$ and query space $\mathcal{X}$. In few-shot in-context learning (ICL), the context $C_k$ is constructed by concatenating $k$ input–label exemplars $(x^{(i)}, y^{(i)})$ formatted via a template function $T$ such that $
C_k = \text{Concat}(T(x^{(1)}, y^{(1)}), \dots, T(x^{(k)}, y^{(k)}))$. Then given the context of $k$-shots and a testing query $x \in \mathcal{X}$, the LLM predicts a label via computing
$$
\hat{y} \in \arg\max_{y \in \mathcal{Y}} P_{\text{LLM}}(y \mid x, C_k).
$$
While ICL offers an appealing alternative to the gradient-based fine-tuning by allowing LLMs to adapt to new tasks via only a handful of in-prompt demonstrations, the resulting posterior distribution $P_{\text{LLM}}(y \mid x,C_k)$ is often distorted by some systematic biases. Such biases inherent in ICL often stem from context examples or their order, which makes $P_{\text{LLM}}(y \mid x,C_k)$
significantly diverge from ground-truth posterior $P^{*}(y|x)$.
Therefore, the objective of calibration  is to refine LLM's predictive probabilities $P_{\text{LLM}}(\cdot \mid x,C_k)$ to align with $P^{*}(y|x)$. 

Existing approaches are mainly focused on correcting the prior distribution of the label via estimating the LLM's internal prior given the context. Despite their successes, one can show that these approaches boil down to merely \text{shifting the LLM's decision boundary}, lacking the ability to alter an LLM's orientation. This limitation turns out to be essential especially in multi-class classification, where an LLM can easily make persistent mistakes. See Figure \ref{fig:calibration_comparison}. Therefore,  to further reduce the biases and align with $P^*(y \mid x)$ in such cases of substantial misorientation, we develop a more principled calibration called Supervised Calibration.

\subsection{Our Proposal}
\label{sec:affine_logit_calibration}

To begin with, we assume the $k$ context examples \( (x^{(i)}, y^{(i)})_{i=1}^k \overset{\text{i.i.d.}}{\sim} P^\ast \). Due to the aforementioned biases, the LLM's posterior \( P_{\text{LLM}}(y \mid x, C_k) \) can deviate notably from the truth \( P^{*}(y \mid x) \). In particular, we measure their deviation via the Kullback–Leibler (KL) divergence defined as
$$
\mathbb{E}_{x \sim P^{*}} \left[ D_{\mathrm{KL}}\big( P^{*}(\cdot \mid x) \,\|\, P_{\text{LLM}}(\cdot \mid x, C_k) \big) \right],
$$
where $D_{\mathrm{KL}}\big(P \,\|\, Q\big) = \sum_{y \in \mathcal{Y}} P(y) \log \frac{P(y)}{Q(y)}$ for some probability measures $P$ and $Q$.
 Let $\Delta^{n}$ be the probability simplex over \( \mathcal{Y} \).  Then to correct for this, we seek a vector-valued calibration function \( f^\ast : \Delta^{n} \to \Delta^{n} \), chosen from a prescribed class \( \mathcal{F} \), such that when applied to the vector of LLM’s predictive probabilities, it minimizes the KL-divergence, i.e.,
\begin{align}
    f^\ast &= \mathop{\arg\min}\limits_{\mathclap{f \in \mathcal{F}}}\text{ }  \mathop{\mathbb{E}}\limits_{\mathclap{x\sim P^{*}}} [ D_{\mathrm{KL}}( P^{*}(\cdot \mid x) \,\|\, f(P_{\mathrm{LLM}}(\cdot | x,C_k)) ) ]\nonumber\\
    &= \mathop{\arg\min}\limits_{\mathclap{f \in \mathcal{F}}}\text{ }-\mathop{\mathbb{E}}\limits_{\mathclap{(x,y)\sim P^{*}}} [ \log(f_y(P_{\mathrm{LLM}}(\cdot | x,C_k))) ],\label{eq: divergence}
\end{align}
where $f_y$ is the $y^{th}$-coordinate projection of $f$. Note that as long as \( \mathcal{F} \) contains the identity map, applying $f^\ast$ enhances the fidelity of 
$P_{\text{LLM}}$. To find $f^\ast$, we highlight two key challenges. 
Firstly, since our method is post-hoc, choosing an effective $\mathcal F$ operating solely on the base LLM predictive probabilities is essential.
Secondly, there is no external data sampled from \( P^\ast \) to approximate the objective function in Equation~\eqref{eq: divergence}. 




\subsubsection{Affine-logit Approximation and Leave-subset-out Strategy}
\label{sec:affine_logit_motivation}


To select an appropriate function class $\mathcal F$, we  only need to consider \( f \) defined over the log-odds of the predictive probabilities against a reference group (class 0 in this paper), since the logistic function is bijective.
Specifically, denote the logits given by the base LLM as
$$
\LogitsM(x; \ContextSet) = \left( \LogitsMc(x; \ContextSet) \triangleq \log \frac{P_{\text{LLM}}(y = c \mid x, C_k)}{P_{\text{LLM}}(y = 0 \mid x, C_k)} \right)_{c = 1}^{n - 1}.
$$
Then, instead, we aim to choose the transformed function class $\widetilde{\mathcal{F}} = \left\{ f : \mathbb{R}^{n-1} \rightarrow \Delta^n \right\}$ for calibration.
To facilitate it, notice that
\begin{align}
P^{*}(y \mid x)
&= \frac{P^{*}(x \mid y) P^{*}(y)}{P^{*}(x)}
\nonumber\\
&\propto P_{\text{LLM}}(y \mid x, C_k) \frac{P^{*}(x \mid y)}{P_{\text{LLM}}(x \mid y, C_k)} \;
  \frac{P^{*}(y)}{P_{\text{LLM}}(y \mid C_k)} \nonumber\\
&\triangleq P_{\text{LLM}}(y \mid x, C_k) \, h(x, y, C_k),
\label{eq:bayes_factor}
\end{align}
which implies that 
\begin{align}\label{eq: shifts}
\TrueLogitsL(x) &= \LogitsMc(x;C_k) +\underbrace{\log\left(\frac{P^{*}(x|c)P_{\text{LLM}}(x|0,C_k)}{P^{*}(x|0)P_{\text{LLM}}(x|c,C_k)}\right)}_{\text{Class Conditional Shift}} \nonumber\\
&+\underbrace{\log\left(\frac{P^{*}(c)P_{\text{LLM}}(0|C_k)}{P^{*}(0)P_{\text{LLM}}(c|C_k)}\right)}_{\text{Label Marginal Shift}}\nonumber\\
&= \LogitsMc(x;C_k) + \log\left(h(x,c,C_k)/ h(x,0,C_k)\right)
\end{align}
where $\TrueLogitsL(x) = \log(P^\ast(c|x) / P^\ast(0 | x))$ is the true logit for class $c$.
Thus, the primary challenge of choosing $\mathcal F$ lies in approximating the unknown correction term
$\log\left(h(x,c,C_k)/ h(x,0,C_k)\right)$ which is ''Class Conditional Shift'' plus ''Label Marginal Shift''.
Since we only have access to the LLM's output logits \( \LogitsM(x; C_k) \), we propose to approximate $\{\TrueLogitsL(x)\}_{c=1}^{n-1}$ via an affine transformation of \( \{m_c(x; C_k)\}_{c=1}^{n-1} \). In particular, our working model $L_c(x; \ParamsIc^k)$ for $c = 1, \dots, n-1$ is
\begin{align}\label{eq:affine_logit_model_formal}
L_c(x; \ParamsIc^k) = w^{k}_c \, \LogitsMc(x; \ContextSet)+b^{k}_c,
\end{align}
where $\ParamsIc^{k} = (b^{k}_c, w^{k}_c)$ are calibration parameters associated with class $c$ and the context size $k$. This affine structure directly targets the two primary sources of discrepancies between true and LLM logits: class-conditional shift and label marginal shift as illustrated in Equation \eqref{eq: shifts}. Specifically, by rearranging Equation \eqref{eq:affine_logit_model_formal} as $$L_c(x; \ParamsIc^k) = \LogitsMc(x; C_k) + [(w^{k}_c-1) \LogitsMc(x; C_k)+b^{k}_c ],$$ we see that the term $(w^{k}_c-1) \LogitsMc(x; C_k)+b^{k}_c$ serves as our learned approximation to the true correction term $\log\left(h(x,c,C_k)/h(x,0,C_k)\right)$.
Within this learned correction, the intercept $b_c^k$ primarily addresses the query-independent "Label Marginal Shift" component from Equation \eqref{eq: shifts}, compensating for discrepancies in label priors. The query-dependent term $(w_c^k-1)\LogitsMc(x;C_k)$ targets the "Class Conditional Shift" by allowing the slope $w_c^k$ to rescale the LLM's original logit $\LogitsMc(x;C_k)$.

\begin{figure}[htbp!]
    \includegraphics[width=1.35\columnwidth]{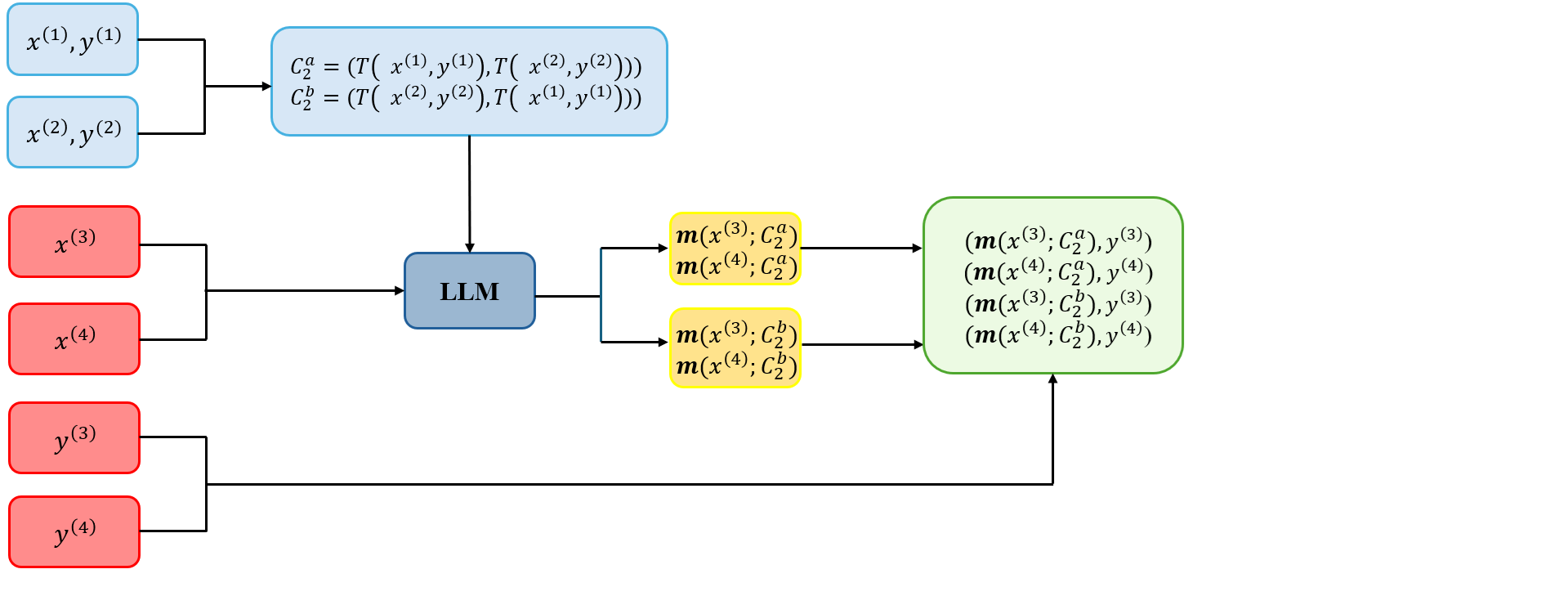}
  \caption{Illustration of surrogate data generation for \( k = 4 \) and \( i = 2 \), depicting two distinct instantiations of the context \( C_2 \subset C_4 \), as described in Algorithm~\ref{alg:ICLGenMLE}. Each context \( C_2^a \) and \( C_2^b \) (shown in blue) is constructed from different orderings of two examples. The remaining examples \( R_2 = C_4 \setminus C_2 \) (shown in red) are passed to the LLM using each context, and the resulting logits \(\LogitsM(x; C_2) \) are paired with their known labels to produce calibration data. Aggregating across such variations enables context-diverse surrogate training.}
\label{fig:surrogate_data}
\end{figure}

Furthermore, $w_c^k$  enables the reorientation of the LLM's decision boundary. For instance, a negative $w_c^k$ inverts the LLM's initial assessment for a class relative to the reference, effectively correcting its predictive direction as illustrated in Figures \ref{fig:calibration_comparison}~(c) and \ref{fig:sst5}. This is a vital capability that methods merely learning a bias (i.e., fixing $w_c^k=1$) lack. As detailed in Section~\ref{sec:theory_combined}, our framework not only unifies but also generalizes several recent ICL calibration techniques. Finally, it naturally encompasses the base LLM's original predictions as a special case when $b^k_c =0$ and $w^k_c=1$ for all $c$. In terms of learning the parameters, if an external calibration dataset $\{(x^{(j)}, y^{(j)})\}_{j=1}^{N_{cal}}$ is provided, we first compute the LLM's logits $\LogitsM(x^{(j)}; C_k)$ for each $x^{(j)}$. Then based on Equation \eqref{eq: divergence}, we estimate the parameters via minimizing the negative log-likelihood, i.e.,
\begin{align}
\label{eq:mle_external}
    \LearnedParams^k &= \arg\min_{\Params^{k}} \{\mathbb L_k(\Params^{k}) \nonumber\\
    &\triangleq -\sum_{j=1}^{N_{cal}} \log f_{y^{(j)}} ( \LogitsM(x^{(j)}; C_k);\Params^{k})\},
\end{align}
where $\Params^{k}=\{\ParamsIc^{k}\}_{c=1}^{n-1}$ and $
f_c(\LogitsM(x^{(j)}; C_k); \Params^{k}) =
\frac{\mathbf{1}_{\{c > 0\}} \exp(L_c(x; \Params^{k}_c)) + \mathbf{1}_{\{c = 0\}}}{1 + \sum_{i=1}^{n-1} \exp(L_i(x; \Params^{k}_i))}
$. This optimization problem is equivalent to standard multi-class logistic regression using the model logits $m_c$ as input features.  However, there is no external calibration dataset available beyond $C_k$. Therefore, we propose generating surrogate training data directly from the demonstration context \( C_k \) via a leave-subset-out strategy. Specifically, we first select a context size $i$ such that $i < k$. We then construct the surrogate training dataset $\mathcal{T}_i$ using Algorithm \ref{alg:ICLGenMLE} in Appendix \ref{append: data_gen_algo and full algo}, as illustrated in Figure \ref{fig:surrogate_data}. Finally, we estimate calibration parameters $\LearnedParams^i$ via minimizing $\mathbb L_i$ under $\mathcal{T}_i$. Note that this method can be applied across multiple context sizes \( i \), enabling ensembling extensions of $\{\LearnedParams^i\}_{i \in I}$ to construct a final estimator for calibration.

\subsubsection{Context Invariance and Directional Trust-Region}\label{sec: context invar}
In the following subsection, we fix the context size \( i  \in \SizeSetI\) and introduce some enhancements to the proposed method. Note that our surrogate data generation process exposes a well-known limitation of ICL, its sensitivity to the composition and ordering of the context. Specifically, a single query pair $(x, y)$ is evaluated using multiple different sub-contexts $C_i$, yielding potentially different logits $\LogitsM(x; C_i)$ and label prediction for the same ground truth label $y$. In essence, an effective calibration method should mitigate this sensitivity, leading to more stable predictions. This motivates incorporating a mechanism to encourage context invariance in the calibrated predictions. To achieve this, we propose augmenting the standard MLE objective (Equation~\eqref{eq:mle_external}) with a context-invariance regularization term.
Specifically, let $C^{(a)}_{i}$ and $C^{(b)}_{i}$ be any two distinct contexts of size $i$ drawn from $C_k$ for evaluating the same query $(x, y)$ in the surrogate data. We aim for the calibrated distributions $\boldsymbol{f}(\LogitsM(x^{(j)}; C^{(a)}_{i}); \boldsymbol{\theta}^i)$ and $\boldsymbol{f}(\LogitsM(x^{(j)}; C^{(b)}_{i}); \boldsymbol{\theta}^i)$, to be similar. To enforce this similarity, we utilize the symmetric cross-entropy between these two calibrated distributions as a regularizer defined as 
\begin{align*}
    &L_{\text{sym}}(\boldsymbol{\theta}^i, x, C^{(a)}_i, C^{(b)}_i)\\
    &=
\mathrm{H}\big(\boldsymbol{f}(\LogitsM(x^{(j)}; C^{(a)}_i); \boldsymbol{\theta}^i),\,
\boldsymbol{f}(\LogitsM(x^{(j)}; C^{(b)}_i); \boldsymbol{\theta}^i)\big),
\end{align*}
where $\mathrm{H}(P, Q) \triangleq -\sum_{c=0}^{n-1} (P_c \log Q_c + Q_c \log P_c)$.
This loss term measures the divergence between the two distributions induced by different contexts, penalizing differences in both directions. Then the overall penalty is defined by averaging $L_{\text{sym}}$ over all possible pairs of contexts associated with each $x$.
\begin{align}
\label{eq:inv_penalty}
\text{InvPenalty}(\boldsymbol{\theta}^i) = \sum_{\substack{\{C^{(a)}_{i},C^{(b)}_{i}\}\\x}} L_{\text{sym}}(\boldsymbol{\theta}^i, x, C^{(a)}_{i},C^{(b)}_{i}).
\end{align}
The full expression of $\text{InvPenalty}$ is given in Equation \eqref{eq:total_inv_penalty_def} of Appendix \ref{append: additional notation}.
On top of ensuring context-invariance, a well-established calibration approach should also take into account the different scenarios induced by the base LLM's reliability and the size of the context. In particular, strong base LLMs warrant minimal adjustment, while weak ones require more aggressive correction, yet limited examples can mislead both cases, risking overfitting or under-correction. To balance this, we regularize the calibration by introducing a \emph{directional trust-region} that restricts parameter updates to remain aligned with the base LLM's logit. Specifically, we constrain the average cosine similarity between each parameter vector $\boldsymbol{\theta}^i_c = [b^i_c, w^i_c]^\top$ and the identity direction $v = [0, 1]^\top$, which corresponds to the base LLM via
$$
\frac{1}{n-1}\sum_{c=1}^{n-1} \frac{(\boldsymbol{\theta}_c^i)^\top v}{\|\boldsymbol{\theta}_c^i\|_2} \ge \tau,
$$
where $\|\cdot \|_2$ refers to $\ell_2$-norm and $\tau \in [0,1]$ modulates the trust: large $\tau$ encourages minor scaling adjustments (exploitation), while smaller values permit broader corrections (exploration). This mirrors trust-region principles in policy optimization (e.g.,  TRPO \citep{schulman2015trust}), adapting model updates based on the confidence in prior predictions.
\subsection{Full Algorithm}\label{sec: full alg}
The final optimization combines this constraint with the likelihood loss and a context-invariance regularizer:
\begin{align}\label{eq:final_opt_problem}
\min_{\boldsymbol{\theta}^i}& \left\{ \sum_{(\LogitsM^{(l)}, y^{(l)}) \in \mathcal{T}_i} \hspace{-1.5em} -\log f_{y^{(l)}} (\LogitsM^{(l)}; \boldsymbol{\theta}^i) + \lambda_{\text{inv}} \text{InvPenalty}(\boldsymbol{\theta}^i) \right\} \; \nonumber\\
&\text{s.t.} \; \frac{1}{n-1} \sum_{c=1}^{n-1} \frac{(\boldsymbol{\theta}_c^i)^\top v}{\|\boldsymbol{\theta}_c^i\|_2} \ge \tau.
\end{align}
where \( \lambda_{\text{inv}}>0 \) is a hyperparameter controlling the strength of the context-invariance penalty. To solve this optimization problem, we used \textit{SciPy's} \texttt{trust-constr} algorithm, a trust-region method designed for constrained optimization.
This optimization can be carried out independently for each \( i \in \SizeSetI \triangleq \{1,\cdots,k-1\}\), resulting in a set of calibration models \( \{\LearnedParams^i\}_{i \in \SizeSetI} \), each specialized for a particular context length. Additionally, at inference, any sub-context \( C_i \) can be used to extract logits for a given size \( i \). This paves the way for a \emph{two-level ensembling strategy} to enhance robustness by aggregating predictions across both multiple context lengths and diverse sub-context samples. Specifically, we train multiple affine-logit models $\{\LearnedParams^i\}_{i \in \SizeSetI}$ using training sets with different sizes of the context. Then, at inference time, given a test query $x_{\text{test}}$, we first
draw $\{C_i^{(j)}\}_{j \in \mathcal{M}_i}$ from $C_k$ for every $i \in I$, where $I$ and $\mathcal M_i$ are user-defined index sets with size $|\mathcal M_i|$ and $|I|$. Then we perform \emph{intra-size} and \emph{inter-size} ensembling by averaging the calibrated predictions over $\{C_i^{(j)}\}_{j \in \mathcal{M}_i}$ and across all context sizes $i\in I$ and output the predictive probability of SC for $x_{\mathrm{test}}$ as
\begin{align}
\hat{\mathbf{p}}_{\mathrm{SC}}(x_{\mathrm{test}})
= \frac{1}{|\SizeSetI|}
\sum_{\substack{i\in \SizeSetI}}\frac{1}{|\mathcal{M}_i|}\left(\sum_{j\in \mathcal{M}_i}\boldsymbol{f}\bigl(\LogitsM(x_{\mathrm{test}};C_i^{(j)});\LearnedParams^{i}\bigr)\right).\nonumber
\end{align}
The final predicted label is 
$
\hat{y}_{\mathrm{SC}}
\in \arg\max_{y_c\in\mathcal{Y}}
[\hat{\mathbf{p}}_{\mathrm{SC}}]_c.$ Overall, this ensembling procedure approximates marginalization over plausible sub-contexts and lengths, significantly improving calibration stability and accuracy. The full algorithm of SC is summarized in Algorithm~\ref{alg:EnsembleAffineLogitCal} of Appendix \ref{append: data_gen_algo and full algo}.

\subsection{Connections to Prior Work and Theoretical Insight}
\label{sec:theory_combined}
In this section, we show the connection of the proposed SC with the existing LM methods and
provide a principled approach to theoretically understand these methods from the perspective of supervised learning.
Specifically, LM methods rely on one core assumption.
\begin{assumption}\label{ass: core}
    The correction term $h(x, y, C_k) \propto \frac{1}{P_{\text{LLM}}(y | C_k)}$.
\end{assumption}
Under Assumption \ref{ass: core}, the derivation in Section \ref{sec:affine_logit_motivation} yields that LM methods are equivalent to assuming
\begin{align}\label{eqn: working model for LM}
    L^\ast_c(x) = \LogitsMc(x; \ContextSet) + B_c(\ContextSet), \text{ } c = 1, \dots, n-1,
\end{align}
where $B_c(\ContextSet) = -\log[P_{\text{LLM}}(c|\ContextSet)/P_{\text{LLM}}(0|\ContextSet)]$.
Therefore, they focus on optimally shifting the decision threshold of the base LLM via estimating $P_{\text{LLM}}(y | C_k)$, which thus gives an estimator for $B_c(\ContextSet)$. We summarize the existing approaches of estimating $P_{\text{LLM}}(y | C_k)$ in  Table~\ref{tab:calibration-formulas} of Appendix \ref{append: additional notation}. However, Assumption \ref{ass: core} can be easily violated in practice, causing model mis-specification error. Therefore, instead of imposing Assumption \ref{ass: core},  we propose to understand existing LM methods from the perspective of function approximation in the supervised learning. In this case, LM methods basically assume a working model \eqref{eqn: working model for LM}. In contrast, the proposed SC considers a strictly larger working model:
\[L_c(x; \ParamsIc^k) = w^{k}_c \, \LogitsMc(x; \ContextSet) + b^{k}_c , \qquad c = 1, \dots, n-1.
\]
This offers a principled framework to compare SC with LM methods and indeed shows that  SC generalizes existing LM methods. Furthermore, within this framework, we analyze these methods via statistical learning theory. Consider a dataset \(\mathcal{T} = \{(x^{(j)}, y^{(j)})\}_{j=1}^N\) of size \(N\), and denote by \(
\hat f:=f_{\LearnedParams^{k}}
\) the solution minimizing \(\mathbb{L}_k(\Params^{k})\) under $\mathcal{T}$. Let \(\mathcal{R}^*\) denote the Bayes risk and \(\mathcal{R}(\hat{f})\) the \(0\text{-}1\) risk of \(\hat{f}\). Then, under standard regularity conditions, the excess risk of SC satisfies, with high probability:
\begin{align}
\label{eq:excess_risk_bound_from_image} 
 \underbrace{\mathcal R(\hat f) - \mathcal R^{\ast}}_{\text{excess risk}}
&\lesssim
\underbrace{\sqrt{D_{\text{KL}}(P^*\,\|\,f^*)-D_{\text{KL}}(P^*\,\|\,P^*)}}_{\text{approximation error}}\nonumber\\
&\;+\;
\sqrt{\frac{2(n-1)}{N}}.
\end{align}
The decomposition leads to the following theoretical insight.  Firstly, thanks to the strictly larger working model, SC attains an approximation error that is guaranteed to be no worse than that of LM methods.
Secondly, SC estimates \( 2(n{-}1) \) parameters—one slope and one intercept per non-reference class—while LM methods estimate only \( n{-}1 \) parameters. This leads to a factor of 2 increase in estimation error, which scales with the number of parameters \( d \) as \( \mathcal{O}(d) \). This gives LM methods an advantage.
However, SC incorporates several variance mitigation strategies to actively control estimation error and fully leverage its lower approximation error: (i) explicit regularization through the directional trust-region constraint and context invariance penalty; and (ii) ensembling procedure in Algorithm~\ref{alg:EnsembleAffineLogitCal}. 

\section{Experiments and Main Results}
In this section, we validate the effectiveness of \MethodName{} by evaluating its classification performance across three LLMs and nine benchmark datasets. \MethodName{} consistently outperforms all baseline calibration methods across various settings, establishing a new state-of-the-art in ICL for classification.

\subsection{Experimental Setup}
\label{subsec:experimental_setup}

\textbf{Datasets.} We evaluate our method on nine text classification benchmarks covering sentiment, topic, and social media analysis: SST-2, SST-5 \citep{socher-etal-2013-recursive}, AG News \citep{zhang2015character}, SUBJ \citep{wang2012baselines}, TREC \citep{li2002learning}, Rotten Tomatoes \citep{pang2005seeingstarsexploitingclass}, TweetEval-Emotion \citep{mohammad-etal-2018-semeval}, TweetEval-Hate \citep{basile2019semeval}, and Financial PhraseBank \citep{malo2014good}.

\textbf{Models and Baselines.} We compare \MethodName{} against the Base LLM and three prior calibration baselines (CC, BC, and DC) on three models: LLaMA-2-7B-Chat-HF \citep{touvron2023llama2openfoundation}, Mistral-7B-Instruct-v0.3\citep{jiang2023mistral7b}, and Qwen2-7B-Instruct \citep{yang2024qwen2technicalreport}. All models are used off-the-shelf from Hugging Face without any fine-tuning. Appendix~\ref{append: implementation details} provides full implementation details for the baselines.

\textbf{Evaluation.} Following prior work, we report Macro-F1 in 4-shot, 8-shot, and 16-shot settings. To ensure robustness, all results are averaged over 5 random seeds on a held-out test set of 256 examples per dataset. Our prompt template is described in Appendix~\ref{append: prompt templates}.

\subsection{Main results}
\begin{figure*}[htbp!]
  \centering
\includegraphics[width=1\textwidth]{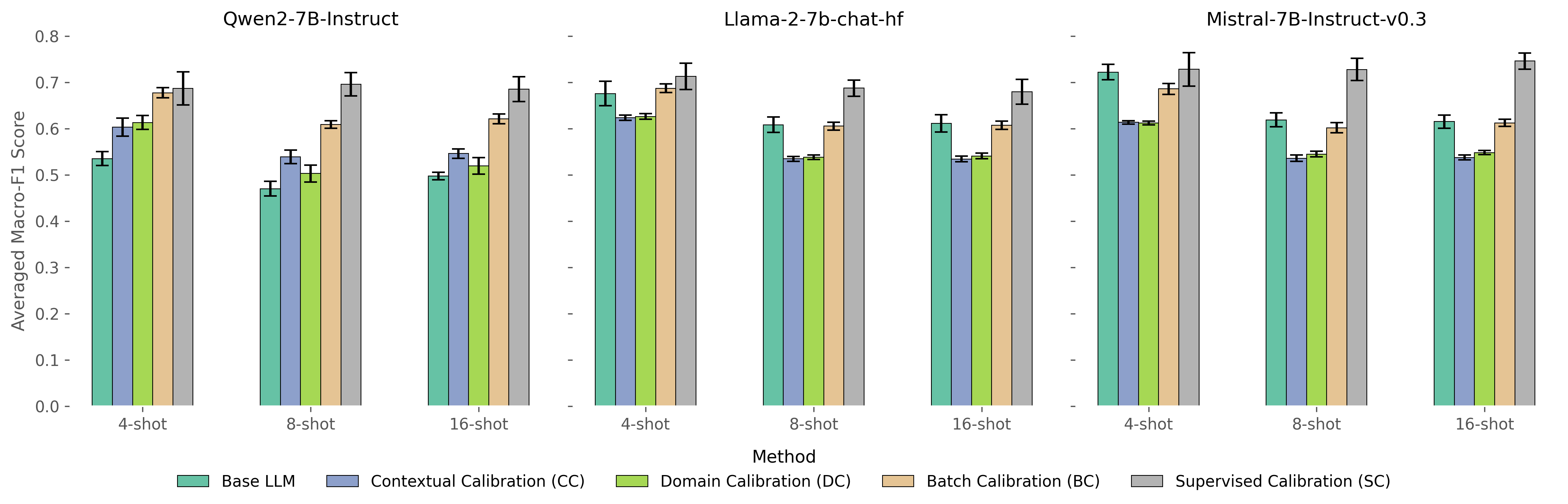}
  \caption{Average Macro-F1 scores for five methods across 9 datasets and 3 LLMs in 4-, 8-, 16-shots settings. Bars show the mean performance and standard deviation across datasets over 5 random seeds.}
  \label{fig:average macro}
\end{figure*}

\begin{figure}[H]
    \centering
    \begin{minipage}{1\columnwidth}
        
        \includegraphics[width=\textwidth]{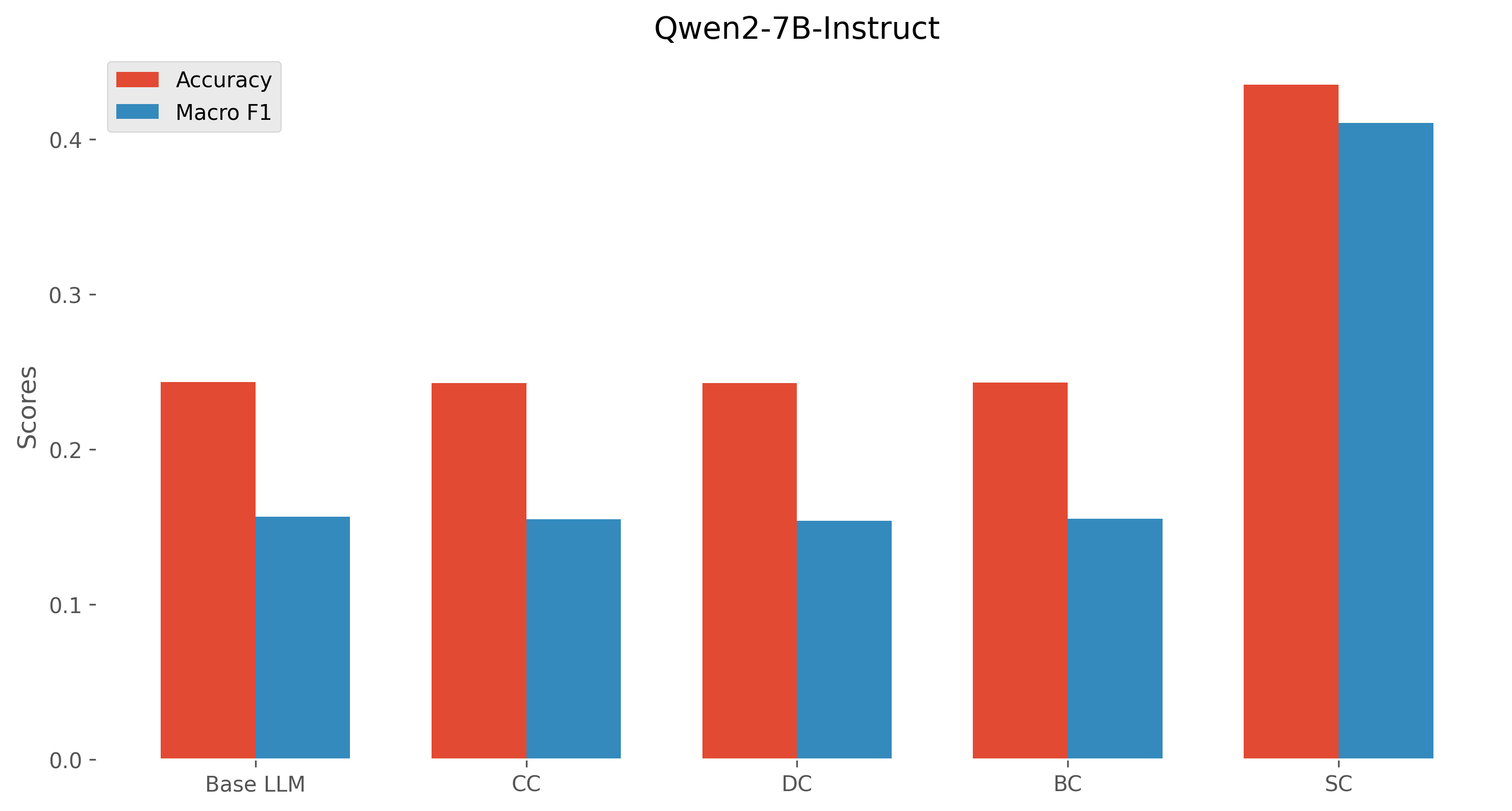}
    \end{minipage}
    \hspace{-1.3cm}
    \begin{minipage}{0.5\textwidth}
        \centering
        \scriptsize
        \begin{tabular}{ccc}
            \toprule
            \textbf{Class} & \textbf{b} & \textbf{w} \\
            \midrule
            Very Negative   (Ref.)      & 0.000 & 0.000 \\
            Negative             & -1.294 & -0.188 \\
            Neutral              & 3.457 & 1.097 \\
         Positive        & 5.541 & 1.190 \\
            Very Positive        & -7.393 & 5.487 \\
            \bottomrule
        \end{tabular}
    \end{minipage}
    \caption{Performance on SST-5 with Qwen2-7B-Instruct in the 8-shot setting, averaged over 5 random seeds. 
    The table on the right shows the average learned coefficients with respect to the \textit{very negative} reference class.}
    \label{fig:sst5}
\end{figure}

Figure~\ref{fig:average macro} reports the Macro-F1 performance of five calibration methods across our full experimental suite (9 datasets, 3 LLMs, 5 seeds, and 3 few-shot settings). Notably, \MethodName{} consistently achieves the highest score across all models and shot counts. In particular compared to the Base LLM, \MethodName{} yields improvements of up to \textbf{+22.6\%} absolute in Macro-F1 (8-shot on Qwen2-7B-Instruct), and on average provides \textbf{+11.1\%} absolute gain across all models and shot configurations. Relative to the strongest competing calibration method (BC), \MethodName{} further improves performance by up to \textbf{+13.4\%} (16-shot on Mistral-7B-Instruct-v0.3) and achieves an average gain of \textbf{+7.1\%}. Overall, these results confirm that \MethodName{} offers a robust and generalizable enhancement of LM methods in few-shot learning. In addition, our numerical results are aligned with our theory presented in Section~\ref{sec:theory_combined}. As shown in Figure~\ref{fig:average macro}, \MethodName{} achieves the highest average score among all methods due to better approximation error, but also exhibits increased variance in its performance.
More detailed numerical results and comparison are given in Appendix \ref{append: tables}.
Furthermore, \MethodName{} delivers a striking improvement on SST-5: in the 8-shot setting with Qwen, it boosts accuracy from 24\% (base LLM) and ~25\% (other methods) to 44\%, nearly doubling performance as shown in Figure~\ref{fig:sst5}. This substantial gain stems from \MethodName{}’s unique ability to not just shift logits, but to reverse the decision boundary when necessary as illustrated in Figure \ref{fig:calibration_comparison}. For instance, it learns a bias of $-1.29$ and a weight of $-0.19$ for the \textit{negative} class relative to \textit{very negative}. This indicates that \MethodName{} effectively shifts and reorients the LLM’s decision boundary between closely related classes, enhancing overall performance.

\paragraph{Ablations.} To isolate the contributions of our framework's key components, we conduct several ablation studies which are detailed in
Appendix~\ref{append: ablation figures}. First, to isolate the impact of learning the per-class scaling factor $w_c$, which underpins SC's ability to reorient decision boundaries, we compare the full SC model against two alternatives in Figure \ref{fig: sc star comb} of Appendix~\ref{app: Scaling Matters}: a restricted variant, $\text{SC}^*$ (where $w_c$ is fixed to 1, thus only learning an optimal bias term), and other baseline calibration methods. Our experiments reveal that $\text{SC}^*$ surpasses these other baselines. This suggests that estimating an optimal bias under \MethodName{} framework is more effective than methods employed by LM methods. More critically, the full SC model achieves higher performance than $\text{SC}^{\ast}$, 
suggesting that the flexibility to learn the scaling factor—and therefore to 
both shift and rescale the LLM's logits—offers a further advantage.

Second, ensembling across context sizes improves performance. In particular, aggregating calibrators trained at heterogeneous context lengths (increasing $|I|$) yields a consistent improvement in both Accuracy and Macro-F1 across multiple LLMs in the 8-shot setting (Figure~\ref{fig: k-shot learners qwen} of Appendix~\ref{app: Ensembling Across}) and similarly in the 16-shot setting (Figure~\ref{fig: 16-shot learners Llama} of Appendix~\ref{app: Ensembling Across}), suggesting that context-size-specific calibrators provide complementary signals.

Third, we investigate the impact of the number of sampled sub-contexts ($m_i$) used for prediction averaging within each context-size-specific calibrator during the ensembling phase. In Figure \ref{fig: Llama icl inference} of Appendix~\ref{eq: sampled sub-contexts}, our findings reveal that increasing $m_i$ (i.e averaging predictions over a greater number of distinct sub-contexts of size $i$) generally enhances Macro-F1 scores. This suggests that more comprehensive sampling of available context variations for each $i$-shot learner improves the accuracy of the ensemble's output.

Fourth, we characterize the computational footprint of sub-context (SC) ensembling by reporting wall-clock training time and inference time per 256 test examples. Timing results show that the added compute from SC comes almost entirely at inference time, and it grows approximately linearly with how many sub-context samples we average over. This is because each additional sampled sub-context requires an additional forward pass through the LLM. In contrast, bias-only baselines are essentially insensitive to this choice and stay nearly flat. Concretely, in the 4-shot setting, SC takes 22.91s per 256 test examples when averaging over 1 sampled sub-context, and increases to 134.96s when averaging over 6 sampled sub-contexts, while baselines remain around 10.5s across all settings (Table~\ref{tab:timing_k4} of Appendix~\ref{eq: Compute and timing}). The same pattern holds for 8-shot: SC increases from 42.83s (1 sampled sub-context) to 260.32s (6 sampled sub-contexts), whereas baselines stay near 11.1s throughout (Table~\ref{tab:timing_k8} of Appendix~\ref{eq: Compute and timing}).

Fifth, we evaluate the performance contributions of our two main components: the directional trust-region constraint and the context invariance penalty. We observe that removing both constraints improves over the uncalibrated baseline, but adding either the directional trust-region constraint or the context-invariance penalty yields further gains, and combining both achieves the strongest performance (Macro-F1 $0.746$ and Acc.\ $0.788$; Table~\ref{tab:ablation} of Appendix~\ref{app: Trust-Region and Invariance}). This suggests that trust-region and invariance are complementary.

Finally, we assess scalability by repeating our evaluation on the larger LLaMA-13B model. Because these runs are substantially more expensive, we focus on three representative datasets—Rotten Tomatoes, SST-2, and AGNews—and compare against the corresponding 7B variant. The results show that SC scales reliably with model size: on LLaMA-13B, SC remains competitive or achieves the best performance on Rotten Tomatoes and SST-2 (Tables~\ref{tab:rotten_tomatoes_13b}--\ref{tab:sst2_13b} of Appendix~\ref{app: Larger Models}). Moreover, the gains persist and can even become more pronounced as the base model grows. For example, on AGNews, SC improves accuracy from $78.12$ with the 7B model to $88.05$ with the 13B model (Table~\ref{tab:agnews_13b} of Appendix~\ref{app: Larger Models}), providing direct evidence that SC’s benefits carry over as underlying model capacity increases.

\section{Conclusion}
In this paper, we introduce \textbf{Supervised Calibration (SC)}, a novel loss-minimization-based calibration framework designed to improve the performance of LLMs in ICL. We design SC to learn a class-specific affine transformation in logit space, allowing it to both shift and reorient the LLM’s decision boundary. Thanks to its expressive functional form, we show that SC generalizes and extends the corrective capabilities of many existing calibration methods for ICL. Looking ahead, several avenues warrant exploration. First, performance could be improved by developing more principled approaches to context selection and weighting, moving beyond the current random sampling strategy. Second, a more rigorous theoretical analysis of SC is needed, particularly one that accounts for the statistical dependencies introduced by our surrogate data generation method. Finally, extending the principles of SC to calibrate LLMs for regression tasks presents a valuable direction for future research.

\section*{Impact Statement}
This paper introduces Supervised Calibration (SC), a calibration framework that improves the reliability of large language models used as few-shot classifiers in in-context learning. By learning a lightweight transformation of the model’s output scores using only the labeled examples already provided in the prompt, SC can reduce systematic prediction biases and make performance less brittle to context composition and ordering. In practical settings where few-shot classification is attractive because collecting large labeled datasets is costly, more stable and better-calibrated predictions can support safer and more dependable downstream pipelines.

At the same time, SC is not a substitute for broader safety and responsible-use practices. Calibrating a model can increase the confidence and apparent consistency of predictions, which may encourage over-reliance if deployed in high-stakes decision-making without appropriate oversight. Moreover, if the in-context labels are biased or unrepresentative, SC may reinforce those biases rather than remove them. We therefore recommend using SC alongside careful dataset and prompt curation, reporting performance variability across contexts, and maintaining human review or fallback mechanisms in sensitive applications. Finally, some SC configurations increase inference-time compute because they average predictions over multiple context variants; practitioners should weigh these accuracy gains against energy considerations.

\nocite{langley00}

\bibliography{example_paper}

@inproceedings{wolf-etal-2020-transformers,
    title = "Transformers: State-of-the-Art Natural Language Processing",
    author = "Wolf, Thomas  and
      Debut, Lysandre  and
      Sanh, Victor  and
      Chaumond, Julien  and
      Delangue, Clement  and
      Moi, Anthony  and
      Cistac, Pierric  and
      Rault, Tim  and
      Louf, Remi  and
      Funtowicz, Morgan  and
      Davison, Joe  and
      Shleifer, Sam  and
      von Platen, Patrick  and
      Ma, Clara  and
      Jernite, Yacine  and
      Plu, Julien  and
      Xu, Canwen  and
      Le Scao, Teven  and
      Gugger, Sylvain  and
      Drame, Mariama  and
      Lhoest, Quentin  and
      Rush, Alexander",
    editor = "Liu, Qun  and
      Schlangen, David",
    booktitle = "Proceedings of the 2020 Conference on Empirical Methods in Natural Language Processing: System Demonstrations",
    month = oct,
    year = "2020",
    address = "Online",
    publisher = "Association for Computational Linguistics",
    url = "https://aclanthology.org/2020.emnlp-demos.6/",
    doi = "10.18653/v1/2020.emnlp-demos.6",
    pages = "38--45"
}

@article{zhou2023batch,
  title={Batch calibration: Rethinking calibration for in-context learning and prompt engineering},
  author={Zhou, Han and Wan, Xingchen and Proleev, Lev and Mincu, Diana and Chen, Jilin and Heller, Katherine and Roy, Subhrajit},
  journal={arXiv preprint arXiv:2309.17249},
  year={2023}
}

@inproceedings{zhao2021calibrate,
  title={Calibrate before use: Improving few-shot performance of language models},
  author={Zhao, Zihao and Wallace, Eric and Feng, Shi and Klein, Dan and Singh, Sameer},
  booktitle={International conference on machine learning},
  pages={12697--12706},
  year={2021},
  organization={PMLR}
}

@misc{brown2020languagemodelsfewshotlearners,
      title={Language Models are Few-Shot Learners}, 
      author={Tom B. Brown and Benjamin Mann and Nick Ryder and Melanie Subbiah and Jared Kaplan and Prafulla Dhariwal and Arvind Neelakantan and Pranav Shyam and Girish Sastry and Amanda Askell and Sandhini Agarwal and Ariel Herbert-Voss and Gretchen Krueger and Tom Henighan and Rewon Child and Aditya Ramesh and Daniel M. Ziegler and Jeffrey Wu and Clemens Winter and Christopher Hesse and Mark Chen and Eric Sigler and Mateusz Litwin and Scott Gray and Benjamin Chess and Jack Clark and Christopher Berner and Sam McCandlish and Alec Radford and Ilya Sutskever and Dario Amodei},
      year={2020},
      eprint={2005.14165},
      archivePrefix={arXiv},
      primaryClass={cs.CL},
      url={https://arxiv.org/abs/2005.14165}, 
}

@misc{liu2021pretrainpromptpredictsystematic,
      title={Pre-train, Prompt, and Predict: A Systematic Survey of Prompting Methods in Natural Language Processing}, 
      author={Pengfei Liu and Weizhe Yuan and Jinlan Fu and Zhengbao Jiang and Hiroaki Hayashi and Graham Neubig},
      year={2021},
      eprint={2107.13586},
      archivePrefix={arXiv},
      primaryClass={cs.CL},
      url={https://arxiv.org/abs/2107.13586}, 
}

@misc{min2022rethinkingroledemonstrationsmakes,
      title={Rethinking the Role of Demonstrations: What Makes In-Context Learning Work?}, 
      author={Sewon Min and Xinxi Lyu and Ari Holtzman and Mikel Artetxe and Mike Lewis and Hannaneh Hajishirzi and Luke Zettlemoyer},
      year={2022},
      eprint={2202.12837},
      archivePrefix={arXiv},
      primaryClass={cs.CL},
      url={https://arxiv.org/abs/2202.12837}, 
}

@inproceedings{qin2024context,
  title={In-context learning with iterative demonstration selection},
  author={Qin, Chengwei and Zhang, Aston and Chen, Chen and Dagar, Anirudh and Ye, Wenming},
  booktitle={Findings of the Association for Computational Linguistics: EMNLP 2024},
  pages={7441--7455},
  year={2024}
}

@article{voronov2024mind,
  title={Mind your format: Towards consistent evaluation of in-context learning improvements},
  author={Voronov, Anton and Wolf, Lena and Ryabinin, Max},
  journal={arXiv preprint arXiv:2401.06766},
  year={2024}
}

@misc{holtzman2022surfaceformcompetitionhighest,
      title={Surface Form Competition: Why the Highest Probability Answer Isn't Always Right}, 
      author={Ari Holtzman and Peter West and Vered Shwartz and Yejin Choi and Luke Zettlemoyer},
      year={2022},
      eprint={2104.08315},
      archivePrefix={arXiv},
      primaryClass={cs.CL},
      url={https://arxiv.org/abs/2104.08315}, 
}

@article{fei2023mitigating,
  title={Mitigating label biases for in-context learning},
  author={Fei, Yu and Hou, Yifan and Chen, Zeming and Bosselut, Antoine},
  journal={arXiv preprint arXiv:2305.19148},
  year={2023}
}

@article{si2023measuring,
  title={Measuring inductive biases of in-context learning with underspecified demonstrations},
  author={Si, Chenglei and Friedman, Dan and Joshi, Nitish and Feng, Shi and Chen, Danqi and He, He},
  journal={arXiv preprint arXiv:2305.13299},
  year={2023}
}

@misc{wang2023largelanguagemodelsfair,
      title={Large Language Models are not Fair Evaluators}, 
      author={Peiyi Wang and Lei Li and Liang Chen and Zefan Cai and Dawei Zhu and Binghuai Lin and Yunbo Cao and Qi Liu and Tianyu Liu and Zhifang Sui},
      year={2023},
      eprint={2305.17926},
      archivePrefix={arXiv},
      primaryClass={cs.CL},
      url={https://arxiv.org/abs/2305.17926}, 
}

@article{pezeshkpour2023sensitivity,
  author  = {Pouya Pezeshkpour and Estevam Hruschka},
  title   = {Large Language Models Sensitivity to the Order of Options in Multiple-Choice Questions},
  journal = {arXiv preprint arXiv:2308.11483},
  year    = {2023}
}

@article{han2022prototypical,
  author  = {Zhixiong Han and Yaru Hao and Li Dong and Yutao Sun and Furu Wei},
  title   = {Prototypical Calibration for Few-Shot Learning of Language Models},
  journal = {arXiv preprint arXiv:2205.10183},
  year    = {2022}
}

@article{cho2024token,
  title={Token-based Decision Criteria Are Suboptimal in In-context Learning},
  author={Cho, Hakaze and Sakai, Yoshihiro and Kato, Mariko and Tanaka, Kenshiro and Ishii, Akira and Inoue, Naoya},
  journal={arXiv preprint arXiv:2406.16535},
  year={2024}
}

@inproceedings{schulman2015trust,
  title={Trust region policy optimization},
  author={Schulman, John and Levine, Sergey and Abbeel, Pieter and Jordan, Michael and Moritz, Philipp},
  booktitle={International conference on machine learning},
  pages={1889--1897},
  year={2015},
  organization={PMLR}
}

@inproceedings{socher-etal-2013-recursive,
    title = "Recursive Deep Models for Semantic Compositionality Over a Sentiment Treebank",
    author = "Socher, Richard  and
      Perelygin, Alex  and
      Wu, Jean  and
      Chuang, Jason  and
      Manning, Christopher D.  and
      Ng, Andrew  and
      Potts, Christopher",
    editor = "Yarowsky, David  and
      Baldwin, Timothy  and
      Korhonen, Anna  and
      Livescu, Karen  and
      Bethard, Steven",
    booktitle = "Proceedings of the 2013 Conference on Empirical Methods in Natural Language Processing",
    month = oct,
    year = "2013",
    address = "Seattle, Washington, USA",
    publisher = "Association for Computational Linguistics",
    url = "https://aclanthology.org/D13-1170/",
    pages = "1631--1642"
}

@article{zhang2015character,
  title={Character-level convolutional networks for text classification},
  author={Zhang, Xiang and Zhao, Junbo and LeCun, Yann},
  journal={Advances in neural information processing systems},
  volume={28},
  year={2015}
}

@inproceedings{wang2012baselines,
  title={Baselines and bigrams: Simple, good sentiment and topic classification},
  author={Wang, Sida I and Manning, Christopher D},
  booktitle={Proceedings of the 50th Annual Meeting of the Association for Computational Linguistics (Volume 2: Short Papers)},
  pages={90--94},
  year={2012}
}

@inproceedings{li2002learning,
  title={Learning question classifiers},
  author={Li, Xin and Roth, Dan},
  booktitle={COLING 2002: The 19th International Conference on Computational Linguistics},
  year={2002}
}

@misc{pang2005seeingstarsexploitingclass,
      title={Seeing stars: Exploiting class relationships for sentiment categorization with respect to rating scales}, 
      author={Bo Pang and Lillian Lee},
      year={2005},
      eprint={cs/0506075},
      archivePrefix={arXiv},
      primaryClass={cs.CL},
      url={https://arxiv.org/abs/cs/0506075}, 
}

@inproceedings{basile2019semeval,
  title={Semeval-2019 task 5: Multilingual detection of hate speech against immigrants and women in twitter},
  author={Basile, Valerio and Bosco, Cristina and Fersini, Elisabetta and Nozza, Debora and Patti, Viviana and Pardo, Francisco Manuel Rangel and Rosso, Paolo and Sanguinetti, Manuela},
  booktitle={Proceedings of the 13th international workshop on semantic evaluation},
  pages={54--63},
  year={2019}
}

@inproceedings{mohammad-etal-2018-semeval,
    title = "{S}em{E}val-2018 Task 1: Affect in Tweets",
    author = "Mohammad, Saif  and
      Bravo-Marquez, Felipe  and
      Salameh, Mohammad  and
      Kiritchenko, Svetlana",
    editor = "Apidianaki, Marianna  and
      Mohammad, Saif M.  and
      May, Jonathan  and
      Shutova, Ekaterina  and
      Bethard, Steven  and
      Carpuat, Marine",
    booktitle = "Proceedings of the 12th International Workshop on Semantic Evaluation",
    month = jun,
    year = "2018",
    address = "New Orleans, Louisiana",
    publisher = "Association for Computational Linguistics",
    url = "https://aclanthology.org/S18-1001/",
    doi = "10.18653/v1/S18-1001",
    pages = "1--17"
}

@article{malo2014good,
  title={Good debt or bad debt: Detecting semantic orientations in economic texts},
  author={Malo, Pekka and Sinha, Ankur and Korhonen, Pekka and Wallenius, Jyrki and Takala, Pyry},
  journal={Journal of the Association for Information Science and Technology},
  volume={65},
  number={4},
  pages={782--796},
  year={2014},
  publisher={Wiley Online Library}
}

@misc{yang2024qwen2technicalreport,
      title={Qwen2 Technical Report}, 
      author={An Yang and Baosong Yang and Binyuan Hui and Bo Zheng and Bowen Yu and Chang Zhou and Chengpeng Li and Chengyuan Li and Dayiheng Liu and Fei Huang and Guanting Dong and Haoran Wei and Huan Lin and Jialong Tang and Jialin Wang and Jian Yang and Jianhong Tu and Jianwei Zhang and Jianxin Ma and Jianxin Yang and Jin Xu and Jingren Zhou and Jinze Bai and Jinzheng He and Junyang Lin and Kai Dang and Keming Lu and Keqin Chen and Kexin Yang and Mei Li and Mingfeng Xue and Na Ni and Pei Zhang and Peng Wang and Ru Peng and Rui Men and Ruize Gao and Runji Lin and Shijie Wang and Shuai Bai and Sinan Tan and Tianhang Zhu and Tianhao Li and Tianyu Liu and Wenbin Ge and Xiaodong Deng and Xiaohuan Zhou and Xingzhang Ren and Xinyu Zhang and Xipin Wei and Xuancheng Ren and Xuejing Liu and Yang Fan and Yang Yao and Yichang Zhang and Yu Wan and Yunfei Chu and Yuqiong Liu and Zeyu Cui and Zhenru Zhang and Zhifang Guo and Zhihao Fan},
      year={2024},
      eprint={2407.10671},
      archivePrefix={arXiv},
      primaryClass={cs.CL},
      url={https://arxiv.org/abs/2407.10671}, 
}

@misc{touvron2023llama2openfoundation,
      title={Llama 2: Open Foundation and Fine-Tuned Chat Models}, 
      author={Hugo Touvron and Louis Martin and Kevin Stone and Peter Albert and Amjad Almahairi and Yasmine Babaei and Nikolay Bashlykov and Soumya Batra and Prajjwal Bhargava and Shruti Bhosale and Dan Bikel and Lukas Blecher and Cristian Canton Ferrer and Moya Chen and Guillem Cucurull and David Esiobu and Jude Fernandes and Jeremy Fu and Wenyin Fu and Brian Fuller and Cynthia Gao and Vedanuj Goswami and Naman Goyal and Anthony Hartshorn and Saghar Hosseini and Rui Hou and Hakan Inan and Marcin Kardas and Viktor Kerkez and Madian Khabsa and Isabel Kloumann and Artem Korenev and Punit Singh Koura and Marie-Anne Lachaux and Thibaut Lavril and Jenya Lee and Diana Liskovich and Yinghai Lu and Yuning Mao and Xavier Martinet and Todor Mihaylov and Pushkar Mishra and Igor Molybog and Yixin Nie and Andrew Poulton and Jeremy Reizenstein and Rashi Rungta and Kalyan Saladi and Alan Schelten and Ruan Silva and Eric Michael Smith and Ranjan Subramanian and Xiaoqing Ellen Tan and Binh Tang and Ross Taylor and Adina Williams and Jian Xiang Kuan and Puxin Xu and Zheng Yan and Iliyan Zarov and Yuchen Zhang and Angela Fan and Melanie Kambadur and Sharan Narang and Aurelien Rodriguez and Robert Stojnic and Sergey Edunov and Thomas Scialom},
      year={2023},
      eprint={2307.09288},
      archivePrefix={arXiv},
      primaryClass={cs.CL},
      url={https://arxiv.org/abs/2307.09288}, 
}

@misc{jiang2023mistral7b,
      title={Mistral 7B}, 
      author={Albert Q. Jiang and Alexandre Sablayrolles and Arthur Mensch and Chris Bamford and Devendra Singh Chaplot and Diego de las Casas and Florian Bressand and Gianna Lengyel and Guillaume Lample and Lucile Saulnier and Lélio Renard Lavaud and Marie-Anne Lachaux and Pierre Stock and Teven Le Scao and Thibaut Lavril and Thomas Wang and Timothée Lacroix and William El Sayed},
      year={2023},
      eprint={2310.06825},
      archivePrefix={arXiv},
      primaryClass={cs.CL},
      url={https://arxiv.org/abs/2310.06825}, 
}
\bibliographystyle{icml2026}

\newpage
\appendix
\onecolumn

\section{Implementation Details}\label{append: implementation details}

\paragraph{Computation Resources.} All large language models (LLMs) used in our experiments are based on publicly available implementations from the \texttt{Hugging Face Transformers} library~\citep{wolf-etal-2020-transformers}. We conduct all experiments on a dedicated computing node equipped with 8 NVIDIA A6000 Ada Generation GPUs.

\paragraph{Contextual Calibration \citep{zhao2021calibrate}(CC)} Following the original CC implementation, we compute the label probabilities conditioned on each of the three content-free tokens—‘N/A’, ‘’, and ‘[MASK]’—along with the context. We then take the mean of these probabilities and use it to normalize the LLM's label-space probabilities computed for the test query and the same context.

\paragraph{Domain-Context Calibration \citep{fei2023mitigating}} We reproduce the DC baseline by using the test set as the unlabeled corpus to construct a bag-of-words. From this bag, we randomly sample tokens to create content-free and in-domain inputs with an average target length. This process is repeated 20 times, and we compute the mean probability over these samples. Following the original implementation, we use this mean to normalize the LLM’s label-space probabilities computed for the test query and context.

\paragraph{Batch Calibration \citep{zhou2023batch} (BC)} BC is an inference-time calibration method that computes the mean of label probabilities over $m$
test samples given the context during the inference. We set $m=128$ and use this mean to normalize the LLM’s label-space probabilities given the test query and context.

\paragraph{Supervised Calibration (\MethodName{})} We adopt an ensembling strategy for \MethodName{} as outlined in Algorithm~\ref{alg:EnsembleAffineLogitCal}. For each configuration—$k = 4$, $k = 8$, and $k = 16$—we set the minimum context size $i_{\text{min}}$ (as defined in Algorithm~\ref{alg:EnsembleAffineLogitCal}) to 1, and the maximum context size $i_{\text{max}}$ to $\min(5, k - 1)$. We fix the regularization parameter $\RegLambda$ to 10 across all settings and LLMs. Additionally, the number of contexts to be sampled from $\mathcal{C}(i)$ (given in Definition \ref{def:Ci}) for size $i$ during the prediction is set as:
\[
m_i = \min\left( \left\lfloor \frac{\mathcal{T}_i}{2} \right\rfloor, 24 \right),
\]
where $\mathcal{T}_i$ denotes the number of available samples for context size $i$.

To determine the value of $\tau$, we use the following formulation:
\[
\tau = \operatorname{cos}(\alpha^{\circ})
\]We first compute the in-sample accuracy of the LLM while generating the training data through Algorithm~\ref{alg:ICLGenMLE}. Based on this accuracy, we set the value of $\alpha^{\circ}$ as follows:
\[
\alpha^{\circ} =
\begin{cases}
20^{\frac{1}{K-1}} & \text{if accuracy} \geq 0.9 \\
45^{\frac{1}{K-1}} & \text{if } 0.7 \leq \text{accuracy} < 0.9 \\
90^{\frac{1}{K-1}} & \text{if } 0.5 \leq \text{accuracy} < 0.7 \\
180 & \text{if accuracy} < 0.5
\end{cases}
\]Here, $K$ denotes the number of distinct labels in the dataset.

While running \MethodName{} with the setting $k=4$, we excluded datasets containing more than four classes (i.e SST5 and TREC). This is because when the number of classes exceeds the number of context examples, some classes are inevitably left out of the training data. This imbalance poses a challenge for training logistic regression models across different context sizes.

\section{Additional Related Work}

\paragraph{Calibration via centroids.} A parallel line of work mitigates in-context biases by replacing the standard decision rule with centroid-based classification. \citet{han2022prototypical} proposed \textbf{Prototypical Calibration}, which models output probability vectors using Gaussian mixtures and assigns labels based on cluster likelihood, improving robustness to prompt variation and class imbalance. Similarly, \citet{cho2024token} introduced \textbf{Hidden Calibration}, which operates in the model’s latent space by computing class centroids over hidden states and classifying based on proximity. Although these methods show empirical performance gains, they rely on additional data beyond the in-context examples, which may not always be available or compatible with the ICL setting.

\section{Prompt Templates}\label{append: prompt templates}
\begin{table}[h!]
\caption{Prompt templates and label words for various datasets.}
\vspace{1em}
\centering
\begin{tabular}{@{}lll@{}}
\toprule
\textbf{Dataset} & \textbf{Prompt Template} & \textbf{Label Words} \\
\midrule
SST2   & sentence: $<$x$>$\textbackslash nsentiment: $<$y$>$ & negative, positive \\
\midrule
SST5   & sentence: $<$x$>$\textbackslash nsentiment: $<$y$>$ & terrible, bad, neutral, good, great \\
\midrule
Rotten T.     & review: $<$x$>$\textbackslash nsentiment: $<$y$>$ & negative, positive \\
\midrule
Financial P.     & sentence: $<$x$>$\textbackslash nsentiment: $<$y$>$ & negative, neutral, positive \\
\midrule
Subj   & review: $<$x$>$\textbackslash ntype: $<$y$>$       & objective, subjective \\
\midrule
TREC   & question: $<$x$>$\textbackslash ntarget: $<$y$>$ & abbreviation, entity, description, person, location, number \\
\midrule
AGNews & news: $<$x$>$\textbackslash ntopic: $<$y$>$       & world, sports, business, technology \\
\midrule
TE-Emo & tweet: $<$x$>$\textbackslash nemotion: $<$y$>$       & anger, joy, optimism, sadness \\
\midrule
TE-Hate & tweet: $<$x$>$\textbackslash nhate speech: $<$y$>$       & non-hate, hate \\
\bottomrule
\end{tabular}\label{tab: prompts}
\end{table}
\section{Additional Notation and Detailed Formulation}\label{append: additional notation}

Let $C_k = \{e^{(1)}, e^{(2)}, \ldots, e^{(k)}\}$ be the full demonstration set of $k$ unique input-label exemplars, where $e^{(l)} = (x^{(l)}, y^{(l)})$. 

\begin{definition}[Set of Ordered Contexts]
\label{def:Ci}
The set $\mathcal{C}(i)$ is defined as:
\begin{align}
\mathcal{C}(i) = \{ (s_1, s_2, \ldots, s_i) \mid s_j \in C_k \quad\text{for}\quad j=1,\ldots,i; \text{ and } s_j \neq s_p \text{ for } j \neq p \}.
\end{align}
This set comprises all distinct ordered sequences (permutations) of $i$ unique exemplars chosen from the full demonstration set $C_k$.
\end{definition}

\begin{definition}[Set of Contexts Used for Query $x$]
\label{def:Cxi}
Given an exemplar $(x,y) \in C_k$, let $\mathcal{T}_i$ be the surrogate training dataset generated by Algorithm 1 using contexts of size $i$ from $C_k$. The set $\mathcal{C}(x, i)$ is defined as:
\begin{align}
\mathcal{C}(x, i) = \{ C^{(j)}_i \in \mathcal{C}(i) \mid (x,y) \notin C^{(j)}_i \text{ and } (m(x; C^{(j)}_i , y) \in \mathcal{T}_i \}.
\end{align}
This set consists of all ordered contexts of size $i$ from $\mathcal{C}(i)$ that do not contain the specific exemplar $(x,y)$ itself, and were actually used to generate a (logit, label) pair for the query $x$ within the surrogate training data $\mathcal{T}_i$.
\end{definition}

\begin{definition}[Context Invariance Regularization Penalty]
\label{def: full_inv_penalty} The total Context Invariance Regularization Penalty for parameters $\boldsymbol{\theta}^i$ is defined as:
\begin{align}
\label{eq:total_inv_penalty_def}
\text{InvPenalty}(\boldsymbol{\theta}^i) = \sum_{x \in \{x_l \mid (x^{(l)}, y^{(l)}) \in C_k\}} \sum_{\{C^{(a)}_{i}, C^{(b)}_{i}\} \subseteq \mathcal{C}(x,i), a \neq b} L_{\text{sym}}(\boldsymbol{\theta}^i, x, C^{(a)}_{i}, C^{(b)}_{i}).
\end{align}
This penalty aggregates the symmetric cross-entropy loss over all distinct pairs of contexts $(C^{(a)}_{i}, C^{(b)}_{i})$ that were used to evaluate each unique query input $x$ derived from the original demonstration set $C_k$. It encourages the calibrated predictions for the same query $x$ to be consistent, regardless of the specific context $C^{(j)}_i \in \mathcal{C}(x,i)$ used to generate the intermediate LLM logits.
\end{definition}
\begin{table}[H]
\centering
\caption{Summary of Label Based calibration methods. Each method adjusts the LLM prediction \( P_{\text{LLM}}(y \mid x, C_k) \) via the different estimators of $P_{\text{LLM}}(y|C_k)$.}
\renewcommand{\arraystretch}{1.4}
\vspace{1em}
\scriptsize
\setlength{\tabcolsep}{4pt}
\rowcolors{2}{gray!5}{white}
\begin{tabular}{>{\centering\arraybackslash}p{3.2cm} 
                >{\centering\arraybackslash}p{6.3cm} 
                >{\raggedright\arraybackslash}p{5.5cm}}
\toprule
\textbf{Method} & \textbf{Formula} & \textbf{Description} \\
\midrule

LLM (Prob) &
$\displaystyle \arg\max_{y} P_{\text{LLM}}(y \mid x, C_k)$ &
Selects the label with the highest conditional probability from the LLM. \\

Contextual Calibration (CC) &
$\displaystyle \arg\max_{y} \frac{P_{\text{LLM}}(y \mid x, C_k)}{P_{\text{LLM}}(y \mid \text{NA}, C_k)}$ &
Normalizes the prediction using a content-free input to reduce label bias. \\

Domain-Context Calibration (DC) &
$\displaystyle \arg\max_{y} \frac{P_{\text{LLM}}(y \mid x, C_k)}{\frac{1}{N} \sum_i P_{\text{LLM}}(y \mid \text{RandDom}_i, C_k)}$ &
Uses randomly sampled domain prompts as a reference for normalization. \\

Batch Calibration (BC) &
$\displaystyle \arg\max_{y} \frac{P_{\text{LLM}}(y \mid x, C_k)}{\frac{1}{N} \sum_i P_{\text{LLM}}(y \mid x_i, C_k)}$ &
Calibrates by averaging predictions over a batch of reference inputs. \\

\bottomrule
\end{tabular}
\label{tab:calibration-formulas}
\end{table}

\section{Full Algorithms}\label{append: data_gen_algo and full algo}
\begin{algorithm}[htbp!]
\caption{Surrogate Data Generation for Calibration}
\label{alg:ICLGenMLE}
\footnotesize
\begin{algorithmic}
\STATE {\bfseries Input:} Demonstration set $C_k = \{(x^{(l)}, y^{(l)})\}_{l=1}^k$ of size $k$;
target context size $i$ such that $1 \le i < k$;
LM inference function $\text{Infer}(x, C)$ that returns logit vector $\mathbf{m}(x;C)$.
\STATE Initialize training set $\mathcal{T}_i \gets \emptyset$.
\STATE Generate $\mathcal{C}(i)$, the set of all distinct ordered subsets of $C_k$ with size $i$
(e.g., permutations of $C_k$, taking first $i$).
\FOR{each context $C_i^{(a)} \in \mathcal{C}(i)$}
    \STATE Define the held-out set $R_i^{(a)} \gets C_k \setminus C_i^{(a)}$ \emph{(set difference based on elements).}
    \FOR{each query $(x,y)$ in $R_i^{(a)}$}
        \STATE Compute model logits: $\mathbf{m}(x;C_i^{(a)}) \gets \text{Infer}(x, C_i^{(a)})$.
        \STATE $\mathcal{T}_i \gets \mathcal{T}_i \cup \{(\mathbf{m}(x;C_i^{(a)}), y)\}$ \emph{(store feature vector and true label).}
    \ENDFOR
\ENDFOR
\STATE {\bfseries Output:} Training set $\mathcal{T}_i$ consisting of pairs (model logits, true label).
\end{algorithmic}
\end{algorithm}

\begin{algorithm}[htbp!]
\caption{\MethodName{} (Full Procedure)}
\label{alg:EnsembleAffineLogitCal}
\footnotesize
\begin{algorithmic}
\STATE {\bfseries Input:} Full demonstration set $C_k = \{(x^{(l)}, y^{(l)})\}_{l=1}^k$;
context sizes $I=\{i_{\min},\dots,i_{\max}\}$;
regularization $\lambda_{\mathrm{inv}}\ge 0$, $\tau\in[0,1]$;
context samples $m_i \ge 1$;
query $x$; inference function $\text{Infer}(x,C)$ returns logit vector $\LogitsM(x,C)$.

\STATE {\bfseries Part 1: Training Phase}
\STATE Initialize parameter set $\Theta \gets \emptyset$.
\FOR{each context size $i \in I$}
    \STATE Generate training data $\mathcal{T}_i$ using Algorithm~\ref{alg:ICLGenMLE} with $C_k$.
    \STATE Learn parameters $\LearnedParams^i$ by solving Eq.~\eqref{eq:final_opt_problem}
           using $\mathcal{T}_i$, $\lambda_{\mathrm{inv}}$, $\tau$.
    \STATE Store $\LearnedParams^i$ in $\Theta$.
\ENDFOR

\STATE {\bfseries Part 2: Prediction Phase (for query $x$)}
\STATE Initialize list $P_{\text{list}} \gets [\,]$.
\FOR{each context size $i \in I$}
    \STATE Sample index set $\mathcal{M}_i \subseteq \{1,\dots,|\mathcal{C}(i)|\}$ uniformly at random
           such that $|\mathcal{M}_i| = m_i$.
    \STATE Retrieve learned parameters $\LearnedParams^i$ from $\Theta$.
    \STATE Retrieve sub-contexts $\{C_i^{(j)}\}_{j \in \mathcal{M}_i}$ from $\mathcal{C}(i)$ using $\mathcal{M}_i$.
    \STATE Initialize list $p_{\text{list}}^{(i)} \gets [\,]$.
    \FOR{each $j \in \mathcal{M}_i$}
        \STATE $\LogitsM(x,C_i^{(j)}) \gets \text{Infer}(x, C_i^{(j)})$.
        \STATE $\mathbf{p}^{(j)}(x) \gets \boldsymbol{f}(\LogitsM(x,C_i^{(j)});\LearnedParams^i)$.
        \STATE Append $\mathbf{p}^{(j)}(x)$ to $p_{\text{list}}^{(i)}$.
    \ENDFOR
    \STATE $\hat{\mathbf{p}}_i(x) \gets \frac{1}{m_i}\sum_{\mathbf{p}(x)\in p_{\text{list}}^{(i)}} \mathbf{p}(x)$.
    \STATE Append $\hat{\mathbf{p}}_i(x)$ to $P_{\text{list}}$.
\ENDFOR
\STATE $\hat{\mathbf{p}}_{\text{\MethodName{}}}(x) \gets
       \frac{1}{|I|}\sum_{\mathbf{p}(x)\in P_{\text{list}}} \mathbf{p}(x)$.
\STATE {\bfseries Output:} $\hat{y}_{\text{\MethodName{}}} \in
\arg\max_{y_c\in\mathcal{Y}} [\hat{\mathbf{p}}_{\text{\MethodName{}}}(x)]_c$.
\end{algorithmic}
\end{algorithm}

\section{Detailed Numerical Results}\label{append: tables}
In this section, we present detailed numerical results. For brevity, we refer to Qwen2-7B-Instruct, Llama-2-7b-chat-hf, and Mistral-7B-Instruct-v0.3 as Qwen, Llama, and Mistral, respectively, throughout the remainder of this section.

\begin{table}[!htbp]
\caption{Average Macro-F1 scores (\%) for various calibration methods on selected datasets, evaluated for each LLM in the 4-shot setting ($k = 4$) over five random seeds. Values are presented as $\text{mean}_{s.d}$, with the highest score in each column highlighted in \textbf{bold} and shaded gray.}

\label{tab:macro_f1_k_4} 
\vspace{1em}
\centering
\scriptsize
\renewcommand{\arraystretch}{1.5}
\setlength{\tabcolsep}{4pt} 
\begin{tabular}{%
    >{\columncolor{white}\centering\arraybackslash}l 
    l 
    ||c|| 
    *{7}{c} 
}
\toprule
Model      & Method    & Avg     & AGNews         & FPB            & SST2           & RT             & Subj            & TE-Emo         & TE-Hate        \\
\midrule
           & Base LLM  & 53.49   & 62.74$_{1.56}$ & 31.22$_{9.82}$ & 87.74$_{7.42}$ & 88.23$_{1.90}$ & 33.02$_{0.81}$ & 35.23$_{1.53}$ & 36.26$_{0.20}$ \\
           & CC        & 60.30   & 85.22$_{4.97}$ & 51.46$_{10.52}$ & \cellcolor{gray!30}\textbf{91.63$_{0.78}$} & 89.91$_{1.35}$ & 38.54$_{7.64}$ & 35.07$_{5.54}$ & 30.25$_{0.00}$ \\
Qwen       & DC        & 61.30   & \cellcolor{gray!30}\textbf{88.68$_{0.68}$} & 52.86$_{10.45}$ & 87.20$_{5.76}$ & \cellcolor{gray!30}\textbf{90.31$_{0.90}$ }& 36.97$_{3.72}$ & \cellcolor{gray!30}\textbf{42.82$_{2.33}$} & 30.25$_{0.00}$ \\
           & BC        & 67.71   & 70.14$_{2.17}$ & 73.54$_{2.75}$ & 88.92$_{5.77}$ & 90.18$_{1.41}$ & \cellcolor{gray!30}\textbf{74.10$_{3.92}$} & 40.94$_{3.24}$ & 36.16$_{0.00}$ \\
           & \MethodName{}        & \cellcolor{gray!30}\textbf{68.66} & 72.76$_{6.13}$ & \cellcolor{gray!30}\textbf{75.57$_{6.67}$} & 90.11$_{4.99}$ & 89.39$_{1.76}$ & 62.23$_{11.15}$ & 41.25$_{17.51}$ & \cellcolor{gray!30}\textbf{49.33$_{8.08}$ } \\
\midrule
           & Base LLM  & 67.57   & \cellcolor{gray!30}\textbf{77.58$_{7.17}$ }& 66.41$_{5.92}$ & 93.36$_{0.44}$ & 91.16$_{1.59}$ & 40.18$_{12.93}$ & \cellcolor{gray!30}\textbf{67.34$_{6.12}$} & 36.94$_{7.64}$ \\
           & CC        & 62.31   & 71.01$_{3.42}$ & 81.86$_{2.72}$ & 93.17$_{1.02}$ & \cellcolor{gray!30}\textbf{92.07$_{0.96}$} & 32.36$_{0.00}$ & 35.45$_{0.76}$ & 30.25$_{0.00}$ \\
Llama      & DC        & 62.61   & 72.10$_{3.61}$ & 82.94$_{2.82}$ & 93.60$_{0.50}$ & 91.95$_{1.18}$ & 32.36$_{0.00}$ & 35.06$_{1.02}$ & 30.25$_{0.00}$ \\
           & BC        & 68.69   & 66.06$_{2.04}$ & \cellcolor{gray!30}\textbf{84.56$_{3.75}$} & 93.53$_{0.47}$ & 91.52$_{1.28}$ & 54.15$_{3.48}$ & 36.29$_{1.38}$ & \cellcolor{gray!30}\textbf{51.70$_{2.00}$ } \\
           & \MethodName{}        & \cellcolor{gray!30}\textbf{71.28 } & 71.76$_{11.31}$ & 84.02$_{4.70}$ & \cellcolor{gray!30}\textbf{94.25$_{0.53}$} & 91.56$_{1.19}$ & \cellcolor{gray!30}\textbf{55.79$_{11.41}$} & 55.35$_{10.57}$ & 46.20$_{4.31}$ \\
\midrule
           & Base LLM  & 72.20   & \cellcolor{gray!30}\textbf{79.28$_{6.90}$} & 89.55$_{1.92}$ & 94.07$_{0.75}$ & 92.47$_{0.62}$ & 35.03$_{6.42}$ & \cellcolor{gray!30}\textbf{60.53$_{9.67}$} & 54.51$_{9.67}$ \\
           & CC        & 61.34   & 63.47$_{1.91}$ & 87.24$_{1.10}$ & 94.76$_{0.70}$ & 92.39$_{0.75}$ & 31.55$_{0.00}$ & 32.11$_{1.24}$ & 27.89$_{0.00}$ \\
Mistral    & DC        & 61.17   & 63.29$_{1.29}$ & 86.08$_{2.53}$ & 94.17$_{0.20}$ & 92.39$_{0.75}$ & 31.55$_{0.00}$ & 32.82$_{1.37}$ & 27.89$_{0.00}$ \\
           & BC        & 68.57   & 62.81$_{1.11}$ & 86.66$_{2.32}$ & 94.00$_{0.69}$ & \cellcolor{gray!30}\textbf{92.63$_{0.67}$} & 48.05$_{6.53}$ & 34.08$_{2.67}$ & \cellcolor{gray!30}\textbf{61.73$_{2.67}$} \\
           & \MethodName{}        & \cellcolor{gray!30}\textbf{72.78} & 75.66$_{11.50}$ & \cellcolor{gray!30}\textbf{90.93$_{2.52}$} & \cellcolor{gray!30}\textbf{95.07$_{1.15}$} & 91.53$_{2.51}$ & \cellcolor{gray!30}\textbf{59.38$_{12.89}$} & 59.48$_{9.90}$ & 37.40$_{16.36}$ \\
\bottomrule
\end{tabular}
\end{table}
\setlength{\aboverulesep}{0.01em} 
\begin{table}[!htbp]
\centering
\caption{Average Macro-F1 scores (\%) for various calibration methods on selected datasets, evaluated for each LLM in the 8-shot setting ($k = 8$) over five random seeds. Values are presented as $\text{mean}_{s.d}$, with the highest score in each column highlighted in \textbf{bold} and shaded gray.}
\label{tab:macro_f1_k_8}
\vspace{1em}
\scriptsize
\renewcommand{\arraystretch}{1.3}
\setlength{\tabcolsep}{4pt}
\begin{tabular}{%
    >{\columncolor{white}\centering\arraybackslash}l 
    l 
    ||c|| 
    *{9}{c} 
}
\toprule
Model      & Method    & Avg     & SST5           & TREC           & AGNews         & FPB            & SST2           & RT             & Subj            & TE-Emo         & TE-Hate        \\
\midrule
           & Base LLM  & 47.00   & 15.65$_{0.33}$ & 45.40$_{5.99}$ & 62.06$_{0.79}$ & 30.13$_{2.09}$ & 74.65$_{18.64}$ & 91.00$_{2.28}$ & 31.55$_{0.00}$ & 34.55$_{2.41}$ & 38.01$_{0.00}$ \\
           & CC        & 53.91   & 15.48$_{0.14}$ & 63.30$_{5.09}$ & 82.27$_{6.74}$ & 35.96$_{7.09}$ & 89.00$_{2.59}$  & \cellcolor{gray!30}\textbf{92.30$_{1.37}$} & 32.67$_{0.96}$ & 46.29$_{5.54}$ & 27.89$_{0.00}$ \\
Qwen       & DC        & 50.26   & 15.41$_{0.07}$ & 43.83$_{3.18}$ & \cellcolor{gray!30}\textbf{86.86$_{0.90}$} & 35.92$_{3.97}$ & 69.94$_{19.04}$ & 91.09$_{1.48}$ & 34.69$_{4.03}$ & 46.74$_{3.82}$ & 27.89$_{0.00}$ \\
           & BC        & 60.88   & 15.52$_{0.12}$ & \cellcolor{gray!30}\textbf{67.98$_{1.73}$} & 65.36$_{1.18}$ & 66.87$_{2.90}$ & 86.43$_{4.45}$  & 91.95$_{1.40}$ & \cellcolor{gray!30}\textbf{76.89$_{1.32}$} & 38.88$_{3.00}$ & 38.01$_{0.00}$ \\
           & \MethodName{}        & \cellcolor{gray!30}\textbf{69.59} & \cellcolor{gray!30}\textbf{41.06$_{2.80}$} & 61.28$_{4.30}$ & 85.32$_{4.37}$ & \cellcolor{gray!30}\textbf{74.97$_{6.19}$} & \cellcolor{gray!30}\textbf{91.36$_{3.75}$} & 90.64$_{2.56}$ & 70.94$_{4.35}$ & \cellcolor{gray!30}\textbf{57.09$_{19.29}$} & \cellcolor{gray!30}\textbf{53.63$_{3.26}$} \\
\midrule
           & Base LLM  & 60.82   & 15.75$_{1.31}$ & \cellcolor{gray!30}\textbf{44.60$_{4.29}$} & 74.55$_{4.43}$ & 80.26$_{2.73}$ & 94.15$_{1.11}$  & 91.94$_{1.17}$ & 37.54$_{5.96}$ & \cellcolor{gray!30}\textbf{68.74$_{3.60}$} & 39.86$_{8.28}$ \\
           & CC        & 53.44   & 30.61$_{1.13}$ & 24.68$_{2.68}$ & 64.66$_{1.50}$ & 80.97$_{2.81}$ & 94.59$_{0.75}$  & 92.40$_{0.72}$ & 31.55$_{0.00}$ & 33.64$_{1.28}$ & 27.89$_{0.00}$ \\
Llama      & DC        & 53.80   & 30.91$_{1.25}$ & 25.52$_{3.12}$ & 65.73$_{0.68}$ & 82.44$_{1.86}$ & 94.47$_{1.29}$  & 92.47$_{0.62}$ & 31.55$_{0.00}$ & 33.25$_{1.10}$ & 27.89$_{0.00}$ \\
           & BC        & 60.52   & 23.49$_{0.80}$ & 36.22$_{1.47}$ & 63.78$_{1.27}$ & 82.71$_{3.05}$ & 94.09$_{1.38}$  & 92.01$_{1.03}$ & \cellcolor{gray!30}\textbf{65.21$_{4.20}$} & 33.56$_{1.15}$ & \cellcolor{gray!30}\textbf{53.59$_{2.51}$} \\
           & \MethodName{}        & \cellcolor{gray!30}\textbf{68.74} & \cellcolor{gray!30}\textbf{42.76$_{4.23}$} & 39.78$_{10.65}$ & \cellcolor{gray!30}\textbf{86.01$_{2.85}$} & \cellcolor{gray!30}\textbf{85.58$_{2.04}$} & \cellcolor{gray!30}\textbf{95.27$_{0.51}$} & \cellcolor{gray!30}\textbf{92.53$_{1.24}$} & 61.89$_{4.20}$ & 66.78$_{5.65}$ & 48.05$_{3.83}$ \\
\midrule
           & Base LLM  & 61.86   & 14.66$_{0.25}$ & 40.08$_{5.39}$ & 70.59$_{3.84}$ & 85.80$_{4.22}$ & 94.41$_{1.75}$  & 92.61$_{0.45}$ & 37.20$_{4.35}$ & 61.82$_{3.01}$ & 59.55$_{6.75}$ \\
           & CC        & 53.70   & 28.22$_{1.26}$ & 27.80$_{3.47}$ & 62.29$_{1.42}$ & 84.64$_{4.39}$ & 94.23$_{1.79}$  & 92.69$_{0.40}$ & 31.55$_{0.00}$ & 32.95$_{1.02}$ & 27.89$_{0.00}$ \\
Mistral    & DC        & 54.47   & 31.15$_{1.38}$ & 30.17$_{3.26}$ & 62.07$_{0.58}$ & 83.59$_{3.07}$ & \cellcolor{gray!30}\textbf{94.68$_{1.56}$} & \cellcolor{gray!30}\textbf{92.70$_{0.46}$} & 31.55$_{0.00}$ & 33.43$_{0.80}$ & 27.89$_{0.00}$ \\
           & BC        & 60.16   & 24.83$_{0.54}$ & 40.26$_{4.25}$ & 61.58$_{0.97}$ & 83.59$_{3.07}$ & 94.19$_{1.52}$  & 92.62$_{0.67}$ & 48.26$_{7.71}$ & 32.91$_{1.05}$ & \cellcolor{gray!30}\textbf{63.25$_{2.06}$} \\
           & \MethodName{}        & \cellcolor{gray!30}\textbf{72.77} & \cellcolor{gray!30}\textbf{45.44$_{3.01}$} & \cellcolor{gray!30}\textbf{48.57$_{8.36}$} & \cellcolor{gray!30}\textbf{86.84$_{3.42}$} & \cellcolor{gray!30}\textbf{88.54$_{4.70}$} & 93.24$_{1.58}$ & 90.09$_{1.73}$ & \cellcolor{gray!30}\textbf{66.91$_{6.13}$} & \cellcolor{gray!30}\textbf{67.73$_{7.99}$} & \cellcolor{gray!30}\textbf{67.53$_{11.74}$} \\
\bottomrule
\end{tabular}
\end{table}

\setlength{\aboverulesep}{0.01em} 
\begin{table}[!htbp]
\centering
\caption{Average Macro-F1 scores (\%) for various calibration methods on selected datasets, evaluated for each LLM in the 16-shot setting ($k = 16$) over five random seeds. Values are presented as $\text{mean}_{s.d}$, with the highest score in each column highlighted in \textbf{bold} and shaded gray.}
\label{tab:macro_f1_k_16}
\vspace{1em}

\scriptsize
\renewcommand{\arraystretch}{1.3}
\setlength{\tabcolsep}{4pt}
\begin{tabular}{%
    >{\columncolor{white}\centering\arraybackslash}l 
    l 
    ||c|| 
    *{9}{c} 
}
\toprule
Model      & Method    & Avg     & SST5           & TREC           & AGNews         & FPB            & SST2           & RT             & Subj            & TE-Emo         & TE-Hate        \\
\midrule
           & Base LLM  & 49.75   & 14.47$_{0.29}$ & 59.68$_{5.52}$ & 63.10$_{0.85}$ & 26.72$_{0.84}$ & 87.55$_{6.49}$ & 91.56$_{1.80}$ & 31.55$_{0.00}$ & 35.15$_{0.56}$ & 38.01$_{0.00}$ \\
           & CC        & 54.57   & 14.41$_{0.21}$ & 69.40$_{1.31}$ & 85.30$_{2.77}$ & 27.16$_{9.25}$ & 92.40$_{0.89}$ & 93.32$_{0.66}$ & 37.69$_{4.80}$ & 43.58$_{0.71}$ & 27.89$_{0.00}$ \\
Qwen       & DC        & 51.92   & 14.38$_{0.21}$ & 44.43$_{3.81}$ & \cellcolor{gray!30}\textbf{88.07$_{0.78}$} & 39.48$_{14.78}$ & 83.91$_{9.82}$ & \cellcolor{gray!30}\textbf{93.42$_{1.05}$} & 35.32$_{4.41}$ & 40.41$_{1.50}$ & 27.89$_{0.00}$ \\
           & BC        & 62.12   & 14.64$_{0.36}$ & \cellcolor{gray!30}\textbf{72.75$_{3.37}$} & 69.02$_{3.35}$ & \cellcolor{gray!30}\textbf{68.42$_{8.43}$} & 91.30$_{0.91}$ & 92.64$_{0.89}$ & \cellcolor{gray!30}\textbf{76.63$_{3.03}$} & 35.63$_{0.92}$ & 38.01$_{0.00}$ \\
           & \MethodName{}       & \cellcolor{gray!30}\textbf{68.52} & \cellcolor{gray!30}\textbf{39.32$_{6.66}$} & 69.91$_{2.56}$ & 85.34$_{3.34}$ & 66.57$_{9.62}$ & \cellcolor{gray!30}\textbf{92.95$_{2.10}$} & 92.15$_{1.39}$ & 66.03$_{10.62}$ & \cellcolor{gray!30}\textbf{53.63$_{6.91}$} & \cellcolor{gray!30}\textbf{50.76$_{10.97}$} \\
\midrule
           & Base LLM  & 60.72   & 14.49$_{0.64}$ & 54.93$_{5.18}$ & 75.64$_{5.72}$ & 76.74$_{5.43}$ & 94.25$_{0.65}$ & 92.01$_{1.17}$ & 37.00$_{4.14}$ & \cellcolor{gray!30}\textbf{69.33$_{9.55}$} & 35.71$_{2.64}$ \\
           & CC        & 53.42   & 31.40$_{1.16}$ & 24.02$_{4.02}$ & 63.73$_{1.29}$ & 81.60$_{2.58}$ & 94.41$_{1.19}$ & \cellcolor{gray!30}\textbf{92.78$_{0.67}$} & 31.55$_{0.00}$ & 33.37$_{1.09}$ & 27.89$_{0.00}$ \\
Llama      & DC        & 54.06   & 32.09$_{1.25}$ & 25.52$_{3.12}$ & 65.54$_{0.68}$ & 83.80$_{3.50}$ & \cellcolor{gray!30}\textbf{94.59$_{1.19}$} & 92.47$_{0.62}$ & 31.55$_{0.00}$ & 32.35$_{1.10}$ & 27.89$_{0.00}$ \\
           & BC        & 60.72   & 24.61$_{1.12}$ & 32.62$_{3.83}$ & 63.85$_{0.57}$ & \cellcolor{gray!30}\textbf{83.37$_{3.68}$} & 94.46$_{0.85}$ & 92.46$_{1.03}$ & \cellcolor{gray!30}\textbf{65.81$_{2.42}$} & 33.64$_{1.28}$ & \cellcolor{gray!30}\textbf{56.26$_{4.20}$} \\
           & \MethodName{}        & \cellcolor{gray!30}\textbf{67.95} & \cellcolor{gray!30}\textbf{42.76$_{4.23}$} & \cellcolor{gray!30}\textbf{62.21$_{5.62}$} & \cellcolor{gray!30}\textbf{87.09$_{2.82}$} & 79.81$_{8.37}$ & 93.81$_{0.71}$ & 91.83$_{1.46}$ & 50.65$_{15.60}$ & 62.21$_{4.15}$ & 46.72$_{10.59}$ \\
\midrule
           & Base LLM  & 61.49   & 14.42$_{0.15}$ & \cellcolor{gray!30}\textbf{45.48$_{4.45}$} & 71.17$_{2.31}$ & 84.17$_{3.03}$ & 93.87$_{0.79}$ & 92.39$_{0.73}$ & 37.69$_{3.27}$ & \cellcolor{gray!30}\textbf{70.79$_{4.21}$} & 43.42$_{9.60}$ \\
           & CC        & 53.75   & 28.96$_{1.12}$ & 28.97$_{3.71}$ & 63.38$_{0.91}$ & 82.73$_{2.58}$ & 93.93$_{0.35}$ & \cellcolor{gray!30}\textbf{92.93$_{0.61}$} & 31.55$_{0.00}$ & 33.39$_{1.09}$ & 27.89$_{0.00}$ \\
Mistral    & DC        & 54.80   & 32.31$_{0.33}$ & 32.79$_{3.07}$ & 62.94$_{0.85}$ & 85.17$_{2.62}$ & \cellcolor{gray!30}\textbf{94.54$_{0.80}$} & 92.15$_{0.49}$ & 31.55$_{0.00}$ & 33.81$_{1.16}$ & 27.89$_{0.00}$ \\
           & BC        & 61.22   & 24.82$_{1.23}$ & 41.11$_{1.87}$ & 63.41$_{0.84}$ & 81.51$_{1.55}$ & 93.40$_{0.58}$ & 92.46$_{0.53}$ & \cellcolor{gray!30}\textbf{56.01$_{6.53}$} & 33.64$_{1.15}$ & \cellcolor{gray!30}\textbf{64.57$_{0.98}$} \\
           & \MethodName{}      & \cellcolor{gray!30}\textbf{74.58} & \cellcolor{gray!30}\textbf{45.92$_{3.25}$} & \cellcolor{gray!30}\textbf{62.50$_{3.97}$} & \cellcolor{gray!30}\textbf{87.42$_{1.83}$} & \cellcolor{gray!30}\textbf{85.98$_{4.47}$} & 94.02$_{1.88}$ & 91.07$_{2.32}$ & \cellcolor{gray!30}\textbf{67.94$_{10.40}$} & \cellcolor{gray!30}\textbf{64.08$_{4.31}$} & \cellcolor{gray!30}\textbf{72.34$_{2.92}$} \\
\bottomrule
\end{tabular}
\end{table}

\begin{table}[!htbp]
\caption{Average Accuracy scores (\%) for various calibration methods on selected datasets, evaluated for each LLM in the 4-shot setting ($k = 4$) over five random seeds. Values are presented as $\text{mean}_{s.d}$, with the highest score in each column highlighted in \textbf{bold} and shaded gray.}
\label{tab:accuracy_k_4}
\vspace{1em}
\centering
\scriptsize
\renewcommand{\arraystretch}{1.3}
\setlength{\tabcolsep}{4pt}
\begin{tabular}{%
    >{\columncolor{white}\centering\arraybackslash}l 
    l 
    ||c|| 
    *{7}{c} 
}
\toprule
Model      & Method    & Avg     & AGNews         & FPB            & SST2           & RT             & Subj            & TE-Emo         & TE-Hate        \\
\midrule
           & Base LLM  & 68.01   & 75.23$_{1.61}$ & 63.36$_{2.91}$ & 87.93$_{7.44}$ & 88.28$_{1.85}$ & 48.16$_{0.38}$  & 56.41$_{1.52}$ & \cellcolor{gray!30}\textbf{56.68$_{0.08}$} \\
           & CC        & 64.34   & 85.47$_{4.80}$ & 50.94$_{13.11}$ & \cellcolor{gray!30}\textbf{91.99$_{0.67}$ }& 89.92$_{1.35}$ & 51.09$_{4.15}$  & 37.58$_{8.27}$ & 43.36$_{0.00}$ \\
Qwen       & DC        & 65.87   & \cellcolor{gray!30}\textbf{88.91$_{0.58}$} & 52.73$_{10.46}$ & 87.30$_{5.81}$ & \cellcolor{gray!30}\textbf{90.31$_{0.90}$} & 50.04$_{1.74}$  & 48.44$_{3.54}$ & 43.36$_{0.00}$ \\
           & BC        & \cellcolor{gray!30}\textbf{74.71 } & 78.28$_{1.53}$ & \cellcolor{gray!30}\textbf{76.64$_{2.85}$} & 89.06$_{3.53}$ & 90.20$_{1.41}$ & \cellcolor{gray!30}\textbf{74.30$_{3.84}$} & \cellcolor{gray!30}\textbf{57.85$_{1.53}$} & 56.64$_{0.00}$ \\
           & \MethodName{}        & 70.62   & 77.34$_{3.89}$ & 74.69$_{9.28}$ & 90.82$_{4.17}$ & 89.41$_{1.74}$ & 65.82$_{8.12}$  & 45.08$_{20.38}$ & 51.17$_{6.97}$ \\
\midrule
           & Base LLM  & 72.86   & \cellcolor{gray!30}\textbf{82.58$_{4.17}$} & 78.55$_{2.69}$ & 93.63$_{0.40}$ & 91.17$_{1.58}$ & 51.88$_{7.02}$  & \cellcolor{gray!30}\textbf{72.85$_{5.23}$} & 46.33$_{3.47}$ \\
           & CC        & 71.40   & 79.30$_{2.02}$ & 85.51$_{2.17}$ & 93.48$_{0.94}$ & \cellcolor{gray!30}\textbf{92.07$_{0.95}$ }& 47.85$_{0.00}$  & 58.20$_{0.92}$  & 43.36$_{0.00}$ \\
Llama      & DC        & 71.47   & 79.61$_{1.89}$ & 85.94$_{2.33}$ & 93.83$_{1.19}$ & 91.95$_{1.18}$ & 47.85$_{0.00}$  & 57.77$_{1.35}$  & 43.36$_{0.00}$ \\
           & BC        & \cellcolor{gray!30}\textbf{74.05 } & 77.19$_{1.27}$ & \cellcolor{gray!30}\textbf{86.99$_{3.09}$ }& 93.75$_{0.48}$ & 91.52$_{1.28}$ & \cellcolor{gray!30}\textbf{58.20$_{3.41}$}  & 58.24$_{1.68}$  & \cellcolor{gray!30}\textbf{52.46$_{1.97}$ } \\
           & \MethodName{}        & 73.78   & 78.12$_{8.67}$ & 86.29$_{2.88}$ & \cellcolor{gray!30}\textbf{94.45$_{0.47}$} & 91.56$_{1.18}$ & 56.45$_{10.80}$ & 61.05$_{14.12}$ & 48.52$_{1.90}$ \\
\midrule
           & Base LLM  & \cellcolor{gray!30}\textbf{76.98} & 82.50$_{4.17}$ & 90.47$_{1.99}$ & 94.22$_{0.76}$ & 92.50$_{0.62}$ & 53.91$_{0.00}$  & 68.36$_{5.16}$  & \cellcolor{gray!30}\textbf{56.88$_{7.72}$} \\
           & CC        & 69.56   & 75.23$_{2.33}$ & 87.34$_{1.57}$ & 94.92$_{0.70}$ & 92.42$_{0.72}$ & 46.09$_{0.00}$  & 52.27$_{2.13}$  & 38.67$_{0.00}$ \\
Mistral    & DC        & 69.44   & 75.31$_{1.81}$ & 85.86$_{3.40}$ & 94.38$_{0.19}$ & 92.42$_{0.72}$ & 46.09$_{0.00}$  & 53.36$_{2.16}$  & 38.67$_{0.00}$ \\
           & BC        & 73.21   & 74.69$_{1.57}$ & 87.19$_{2.28}$ & 94.14$_{0.70}$ & \cellcolor{gray!30}\textbf{92.66$_{0.67}$} & 48.44$_{9.55}$  & 53.13$_{1.40}$  & \cellcolor{gray!30}\textbf{62.27$_{2.35}$} \\
           & \MethodName{}        & 75.59   & \cellcolor{gray!30}\textbf{80.23$_{9.04}$} & \cellcolor{gray!30}\textbf{92.50$_{1.87}$} & \cellcolor{gray!30}\textbf{95.23$_{1.09}$} & 91.56$_{1.45}$ & \cellcolor{gray!30}\textbf{62.03$_{8.98}$} & \cellcolor{gray!30}\textbf{65.23$_{14.12}$} & 42.27$_{17.35}$ \\
\bottomrule
\end{tabular}
\end{table}

\begin{table}[!htbp]
\centering
\caption{Average Accuracy scores (\%) for various calibration methods on selected datasets, evaluated for each LLM in the 8-shot setting ($k = 8$) over five random seeds. Values are presented as $\text{mean}_{s.d}$, with the highest score in each column highlighted in \textbf{bold} and shaded gray.}
\label{tab:accuracy_k_8}
\vspace{1em}

\scriptsize
\renewcommand{\arraystretch}{1.3}
\setlength{\tabcolsep}{4pt}
\begin{tabular}{%
    >{\columncolor{white}\centering\arraybackslash}l  
    l                                                
    ||c||                                            
    *{9}{c}                                          
}
\toprule
Model      & Method    & Avg     & SST5           & TREC           & AGNews         & FPB            & SST2           & RT             & Subj            & TE-Emo         & TE-Hate        \\
\midrule
           & Base LLM  & 60.32   & 24.34$_{0.16}$ & 54.22$_{7.21}$ & 73.16$_{0.81}$ & 62.58$_{0.52}$ & 76.09$_{16.30}$ & 91.02$_{2.26}$ & 46.09$_{0.00}$ & 54.06$_{2.74}$ & 61.33$_{0.00}$ \\
           & CC        & 58.59   & 24.26$_{0.08}$ & 67.34$_{3.93}$ & 83.98$_{4.99}$ & 33.28$_{6.70}$ & 89.53$_{2.29}$  & 92.30$_{1.37}$ & 46.64$_{0.47}$ & 51.33$_{7.54}$ & 38.67$_{0.00}$ \\
Qwen       & DC        & 55.94   & 24.26$_{0.08}$ & 52.81$_{1.70}$ & \cellcolor{gray!30}\textbf{87.27$_{0.79}$} & 33.13$_{4.34}$ & 71.72$_{16.75}$ & 91.09$_{1.48}$ & 47.66$_{2.05}$ & 56.88$_{3.25}$ & 38.67$_{0.00}$ \\
           & BC        & 68.64   & 24.30$_{0.10}$ & \cellcolor{gray!30}\textbf{73.59$_{1.79}$} & 74.65$_{0.69}$ & 70.70$_{3.10}$ & 86.52$_{4.48}$  & \cellcolor{gray!30}\textbf{91.95$_{1.40}$} & \cellcolor{gray!30}\textbf{77.03$_{1.34}$} & 57.66$_{1.63}$ & \cellcolor{gray!30}\textbf{61.33$_{0.00}$} \\
           & \MethodName{}        & \cellcolor{gray!30}\textbf{72.30} & \cellcolor{gray!30}\textbf{43.52$_{4.28}$} & 69.06$_{1.32}$ & 86.02$_{4.01}$ & \cellcolor{gray!30}\textbf{76.33$_{6.70}$} & \cellcolor{gray!30}\textbf{91.88$_{3.27}$} & 90.70$_{2.46}$ & 72.50$_{3.45}$ & \cellcolor{gray!30}\textbf{61.33$_{20.50}$} & 59.38$_{3.81}$ \\
\midrule
           & Base LLM  & 66.62   & 23.12$_{0.67}$ & \cellcolor{gray!30}\textbf{56.88$_{4.73}$} & 80.08$_{2.99}$ & 84.14$_{3.44}$ & 94.30$_{1.12}$  & 91.95$_{1.17}$ & 48.75$_{3.30}$ & \cellcolor{gray!30}\textbf{75.39$_{3.59}$} & 45.00$_{5.04}$ \\
           & CC        & 63.59   & \cellcolor{gray!30}\textbf{50.08$_{3.00}$} & 36.48$_{3.58}$ & 76.09$_{1.24}$ & 82.73$_{3.44}$ & 94.77$_{0.72}$  & 92.42$_{0.72}$ & 46.09$_{0.00}$ & 55.00$_{2.06}$ & 38.67$_{0.00}$ \\
Llama      & DC        & 63.71   & 48.59$_{3.28}$ & 37.66$_{3.71}$ & 76.80$_{0.88}$ & 84.06$_{2.90}$ & 94.61$_{1.29}$  & \cellcolor{gray!30}\textbf{92.50$_{0.63}$} & 46.09$_{0.00}$ & 54.37$_{1.86}$ & 38.67$_{0.00}$ \\
           & BC        & 66.55   & 31.48$_{1.37}$ & 47.50$_{1.69}$ & 75.78$_{1.64}$ & 83.44$_{3.69}$ & 94.22$_{1.38}$  & 92.03$_{1.04}$ & \cellcolor{gray!30}\textbf{65.78$_{4.30}$} & 54.77$_{1.70}$ & \cellcolor{gray!30}\textbf{53.91$_{2.60}$} \\
           & \MethodName{}        & \cellcolor{gray!30}\textbf{71.61} & 45.94$_{5.52}$ & 50.86$_{10.44}$ & \cellcolor{gray!30}\textbf{86.56$_{2.76}$} & \cellcolor{gray!30}\textbf{86.95$_{2.35}$} & \cellcolor{gray!30}\textbf{95.39$_{0.52}$} & 92.58$_{1.24}$ & 63.05$_{3.39}$ & 73.52$_{4.08}$ & 49.61$_{3.55}$ \\
\midrule
           & Base LLM  & 68.27   & 23.05$_{0.25}$ & 51.48$_{5.31}$ & 76.88$_{1.66}$ & 85.86$_{5.64}$ & 94.53$_{1.75}$  & 92.66$_{0.46}$ & 54.84$_{1.88}$ & 74.30$_{1.84}$ & 60.86$_{5.49}$ \\
           & CC        & 64.42   & \cellcolor{gray!30}\textbf{54.22$_{0.52}$} & 40.86$_{3.51}$ & 74.45$_{1.67}$ & 84.61$_{5.70}$ & 94.38$_{1.76}$  & 92.73$_{0.40}$ & 46.09$_{0.00}$ & 53.75$_{1.69}$ & 38.67$_{0.00}$ \\
Mistral    & DC        & 65.02   & 54.06$_{0.72}$ & 43.36$_{3.44}$ & 74.30$_{0.62}$ & \cellcolor{gray!30}\textbf{86.64$_{5.28}$} & \cellcolor{gray!30}\textbf{94.84$_{1.51}$} & 92.73$_{0.47}$ & 46.09$_{0.00}$ & 54.45$_{1.25}$ & 38.67$_{0.00}$ \\
           & BC        & 66.35   & 34.92$_{0.53}$ & \cellcolor{gray!30}\textbf{51.48$_{4.13}$} & 73.67$_{1.12}$ & 83.91$_{4.18}$ & 94.30$_{1.51}$  & 92.66$_{0.67}$ & 48.75$_{7.86}$ & 53.67$_{1.76}$ & 63.83$_{1.86}$ \\
           & \MethodName{}        & \cellcolor{gray!30}\textbf{75.54} & 48.52$_{5.84}$ & 57.58$_{8.88}$ & \cellcolor{gray!30}\textbf{87.42$_{3.19}$} & \cellcolor{gray!30}\textbf{89.53$_{5.11}$} & 93.59$_{1.41}$ & \cellcolor{gray!30}\textbf{90.16$_{1.68}$} & \cellcolor{gray!30}\textbf{67.50$_{6.08}$} & \cellcolor{gray!30}\textbf{75.16$_{5.08}$} & \cellcolor{gray!30}\textbf{70.39$_{6.96}$} \\
\bottomrule
\end{tabular}
\end{table}


\begin{table}[!htbp]
\centering
\caption{Average Accuracy scores (\%) for various calibration methods on selected datasets, evaluated for each LLM in the 16-shot setting ($k = 16$) over five random seeds. Values are presented as $\text{mean}_{s.d}$, with the highest score in each column highlighted in \textbf{bold} and shaded gray.}
\label{tab:accuracy_k_16}
\vspace{1em}

\scriptsize
\renewcommand{\arraystretch}{1.3}
\setlength{\tabcolsep}{4pt}
\begin{tabular}{%
    >{\columncolor{white}\centering\arraybackslash}l 
    l 
    ||c|| 
    *{9}{c} 
}
\toprule
Model      & Method    & Avg     & SST5           & TREC           & AGNews         & FPB            & SST2           & RT             & Subj            & TE-Emo         & TE-Hate        \\
\midrule
           & Base LLM  & 62.71   & 22.81$_{0.19}$ & 60.16$_{3.27}$ & 75.62$_{0.94}$ & 64.06$_{0.00}$ & 87.66$_{6.53}$ & 91.56$_{1.81}$ & 46.09$_{0.00}$ & 55.08$_{0.96}$ & \cellcolor{gray!30}\textbf{61.33$_{0.00}$} \\
           & CC        & 59.08   & 22.81$_{0.19}$ & 70.00$_{1.29}$ & 86.17$_{2.18}$ & 25.08$_{8.06}$ & 92.66$_{0.83}$ & 93.36$_{0.65}$ & 49.30$_{2.61}$ & 53.67$_{1.99}$ & 38.67$_{0.00}$ \\
Qwen       & DC        & 57.47   & 22.81$_{0.19}$ & 51.02$_{3.73}$ & \cellcolor{gray!30}\textbf{88.44$_{0.72}$} & 37.03$_{15.89}$ & 84.14$_{9.51}$ & \cellcolor{gray!30}\textbf{93.44$_{1.06}$} & 48.05$_{2.37}$ & 53.59$_{1.09}$ & 38.67$_{0.00}$ \\
           & BC        & 69.49   & 22.89$_{0.19}$ & 73.91$_{2.07}$ & 78.05$_{1.59}$ & \cellcolor{gray!30}\textbf{72.81$_{8.05}$} & 91.41$_{0.92}$ & 92.66$_{0.90}$ & \cellcolor{gray!30}\textbf{76.80$_{2.87}$} & 55.55$_{0.62}$ & 61.33$_{0.00}$ \\
           & \MethodName{}        & \cellcolor{gray!30}\textbf{70.77}   & \cellcolor{gray!30}\textbf{41.64$_{6.65}$} & \cellcolor{gray!30}\textbf{73.98$_{2.67}$} & 85.78$_{3.38}$ & 67.11$_{11.83}$ & \cellcolor{gray!30}\textbf{93.20$_{1.89}$} & 92.19$_{1.42}$ & 68.52$_{8.27}$ & \cellcolor{gray!30}\textbf{60.23$_{5.91}$} & 54.30$_{8.05}$ \\
\midrule
           & Base LLM  & 66.97   & 22.50$_{0.31}$ & \cellcolor{gray!30}\textbf{65.86$_{3.44}$} & 81.09$_{2.71}$ & 81.72$_{5.09}$ & 94.45$_{0.57}$ & 92.03$_{1.27}$ & 48.75$_{2.09}$ & \cellcolor{gray!30}\textbf{73.20$_{6.24}$} & 43.12$_{3.08}$ \\
           & CC        & 63.88   & \cellcolor{gray!30}\textbf{52.58$_{0.80}$} & 37.19$_{3.21}$ & 75.86$_{1.47}$ & 82.34$_{5.05}$ & 94.61$_{1.09}$ & \cellcolor{gray!30}\textbf{92.81$_{0.68}$} & 46.09$_{0.00}$ & 54.77$_{0.76}$ & 38.67$_{0.00}$ \\
Llama      & DC        & 64.11   & 50.70$_{1.84}$ & 39.14$_{4.17}$ & 76.95$_{0.86}$ & \cellcolor{gray!30}\textbf{85.23$_{4.79}$} & \cellcolor{gray!30}\textbf{94.77$_{1.15}$} & 92.81$_{0.80}$ & 46.09$_{0.00}$ & 52.66$_{1.70}$ & 38.67$_{0.00}$ \\
           & BC        & 66.86   & 33.36$_{1.32}$ & 44.22$_{3.30}$ & 76.25$_{0.80}$ & 83.52$_{5.10}$ & 94.61$_{0.83}$ & 92.34$_{1.12}$ & \cellcolor{gray!30}\textbf{66.64$_{2.38}$} & 54.30$_{0.35}$ & \cellcolor{gray!30}\textbf{56.48$_{4.29}$} \\
           & \MethodName{}        & \cellcolor{gray!30}\textbf{70.92}   & 44.45$_{4.58}$ & 65.47$_{4.60}$ & \cellcolor{gray!30}\textbf{87.42$_{2.90}$} & 78.83$_{12.25}$ & 93.98$_{0.77}$ & 91.88$_{1.45}$ & 56.88$_{8.83}$ & 66.64$_{5.59}$ & 52.73$_{11.41}$ \\
\midrule
           & Base LLM  & 67.65   & 22.73$_{0.16}$ & 56.88$_{4.99}$ & 78.67$_{1.92}$ & 83.91$_{3.73}$ & 93.98$_{0.80}$ & 92.42$_{0.72}$ & 55.08$_{1.44}$ & \cellcolor{gray!30}\textbf{76.48$_{3.62}$} & 48.67$_{6.38}$ \\
           & CC        & 64.57   & \cellcolor{gray!30}\textbf{54.30$_{0.55}$} & 42.66$_{4.30}$ & 75.78$_{1.05}$ & 82.03$_{3.38}$ & 94.06$_{0.38}$ & \cellcolor{gray!30}\textbf{92.97$_{0.61}$} & 46.09$_{0.00}$ & 54.53$_{1.74}$ & 38.67$_{0.00}$ \\
Mistral    & DC        & 65.41   & 54.14$_{0.72}$ & 47.11$_{4.27}$ & 75.47$_{1.03}$ & 85.08$_{3.46}$ & \cellcolor{gray!30}\textbf{94.69$_{0.80}$} & 92.19$_{0.49}$ & 46.09$_{0.00}$ & 55.23$_{1.76}$ & 38.67$_{0.00}$ \\
           & BC        & 67.52   & 34.69$_{1.32}$ & 53.28$_{2.16}$ & 75.94$_{1.01}$ & 81.25$_{2.14}$ & 93.52$_{0.58}$ & 92.50$_{0.52}$ & 56.56$_{6.32}$ & 55.00$_{1.81}$ & 64.92$_{0.90}$ \\
           & \MethodName{}        & \cellcolor{gray!30}\textbf{76.96}   & 47.27$_{2.43}$ & \cellcolor{gray!30}\textbf{73.28$_{3.01}$} & \cellcolor{gray!30}\textbf{87.81$_{1.81}$} & \cellcolor{gray!30}\textbf{85.78$_{6.80}$} & 94.30$_{1.70}$ & 91.09$_{2.34}$ & \cellcolor{gray!30}\textbf{70.86$_{7.29}$} & 68.05$_{5.05}$ & \cellcolor{gray!30}\textbf{74.22$_{1.38}$} \\
\bottomrule
\end{tabular}
\end{table}

\section{Ablation Results}\label{append: ablation figures}
We conduct ablation studies to dissect the distinct contributions of key components within our Supervised Calibration (SC) framework.

\subsection{Scaling Matters}\label{app: Scaling Matters}

\begin{figure}[h]
    \centering 
    \includegraphics[width=0.8\columnwidth]{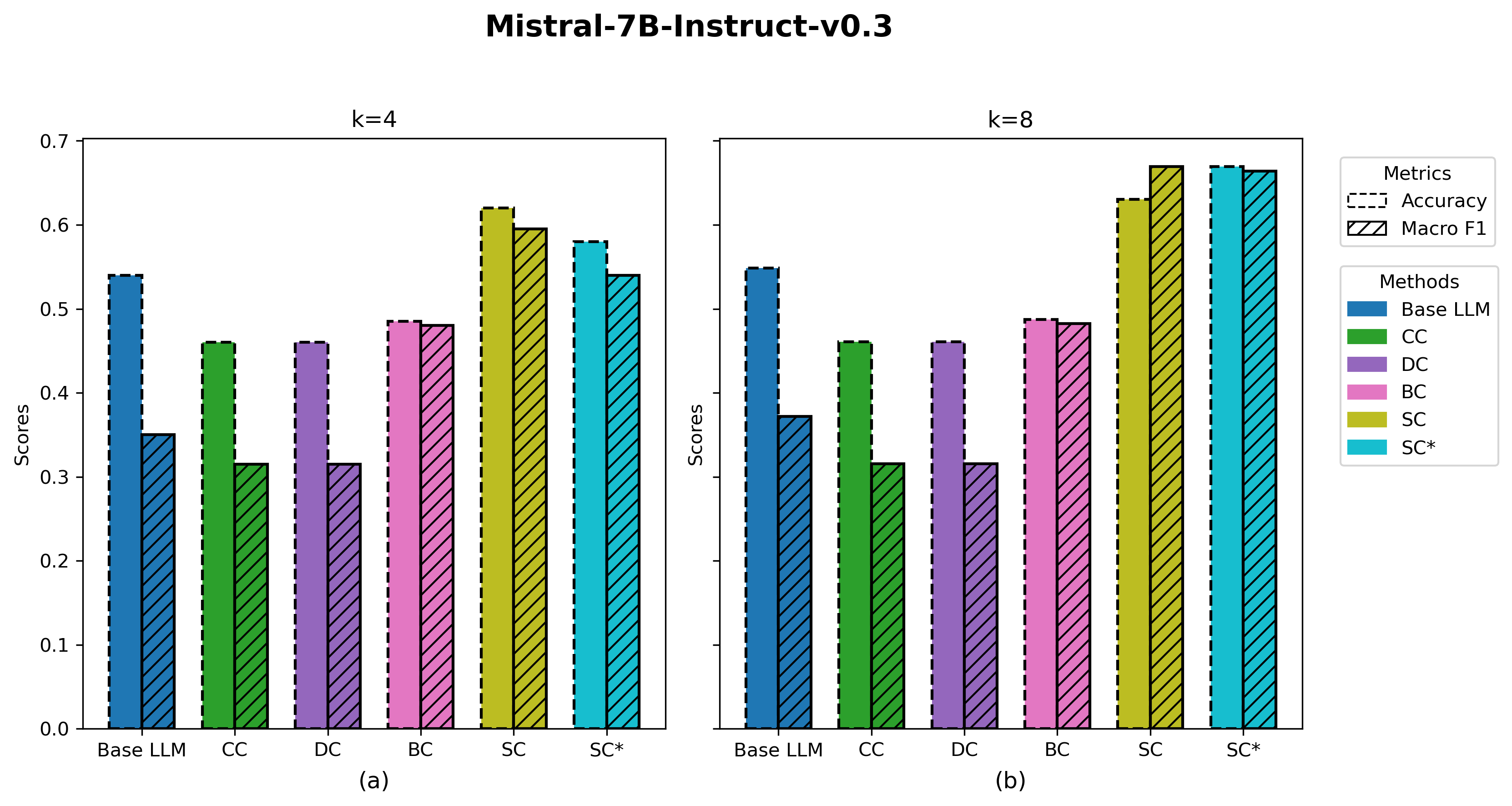}
    \caption{Accuracy and Macro-F1 scores of six methods on the Subjective dataset using the Mistral-7B-Instruct-v0.3 model in (a) 4-shot and (b) 8-shot settings. Results are averaged over 5 random seeds. Bars represent the mean performance for each metric as indicated in the legend. $\text{SC}^*$ stands for the case where the scaling factor $w_c$ is fixed to 1 under the SC framework. Notably, SC consistently outperforms all other methods in both settings. The improved performance of $\text{SC}^*$ over other baselines suggests that estimating an optimal bias under SC framework is more effective than the methods employed by LM approaches, while the full SC further demonstrates the advantage of also learning the scaling factor.}
    \label{fig: sc star comb}
\end{figure}

First, to isolate the impact of learning the per-class scaling factor $w_c$, which underpins SC's ability to reorient decision boundaries, we compare the full SC model against two alternatives in Figure \ref{fig: sc star comb}: a restricted variant, $\text{SC}^*$ (where $w_c$ is fixed to 1, thus only learning an optimal bias term), and other baseline calibration methods. Our experiments reveal that $\text{SC}^*$ surpasses these other baselines. This suggests that estimating an optimal bias under \MethodName{} framework is more effective than methods employed by LM methods. More critically, the full SC model achieves higher performance than $\text{SC}^{\ast}$, 
suggesting that the flexibility to learn the scaling factor—and therefore to 
both shift and rescale the LLM's logits—offers a further advantage. 

The performance difference between \MethodName{} and $\text{SC}^{\ast}$ is particularly apparent on a challenging 8-shot, multi-class classification task (SST-5) where the base model's predictions are often poorly oriented. Specifically, Table \ref{tab:hard_task} shows that $\text{SC}^{\ast}$ method achieves a very low Macro-F1 of 0.1004, indicating its inability to correct the model's predictions. In stark contrast, the full \MethodName{} method boosts the Macro-F1 to 0.4106 and accuracy to 0.4352, representing a four-fold improvement. This vast performance gap confirms our hypothesis: on difficult tasks with severe miscalibrations, only full \MethodName{}, capable of both shifting and scaling the decision boundary, can effectively correct severly misaligned LLM.

\begin{table}[h!]
\centering
\caption{Comparison on the 8-shot SST-5 task with the Qwen2-7B-Instruct model. SC v.s SC*.}
\label{tab:hard_task}
\begin{tabular}{lcc}
\toprule
\textbf{Method} & \textbf{Macro-F1 (mean $\pm$ SE)} & \textbf{Accuracy (mean $\pm$ SE)} \\
\midrule
Base LLM & $0.1565 \pm 0.0033$ & $0.2434 \pm 0.0016$ \\
SC* (scaling=1) & $0.1004 \pm 0.0125$ & $0.2227 \pm 0.0168$ \\
SC              & $0.4106 \pm 0.0280$ & $0.4352 \pm 0.0428$ \\
\bottomrule
\end{tabular}
\end{table}

\subsection{Ensembling Across Context Sizes (\(|I|\)) Improves Performance}\label{app: Ensembling Across}
\begin{figure}[h]
    
    \includegraphics[width=\columnwidth]{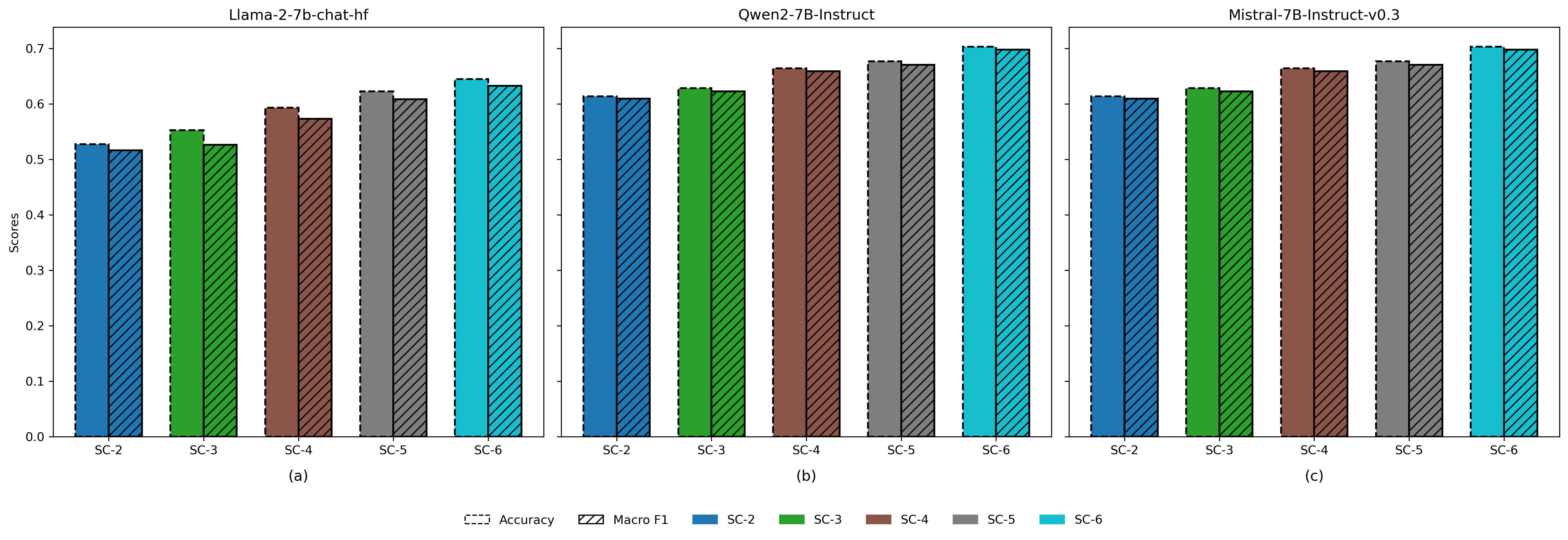}
    \caption{Impact of ensembling context-size-specific models within the \MethodName{} framework on the Subjective dataset in an 8-shot setting. Results are reported for (a) Llama-2-7b-chat-hf, (b) Qwen2-7B-Instruct, and (c) Mistral-7B-Instruct-v0.3, using Accuracy and Macro-F1 scores averaged over 5 random seeds. Each ensemble, denoted SC-\textit{N}, aggregates calibration models trained on context sizes ranging from 1 to $N$ (e.g., SC-2 uses models with context sizes 1 and 2, SC-6 includes context sizes 1 through 6). The consistent improvement in performance as N increases across all three LLMs highlights the general benefit of aggregating insights from a more diverse set of k-shot learners.}
    \label{fig: k-shot learners qwen}
\end{figure}

\begin{figure}[h]
    \centering 
    \includegraphics[width=0.5\columnwidth]{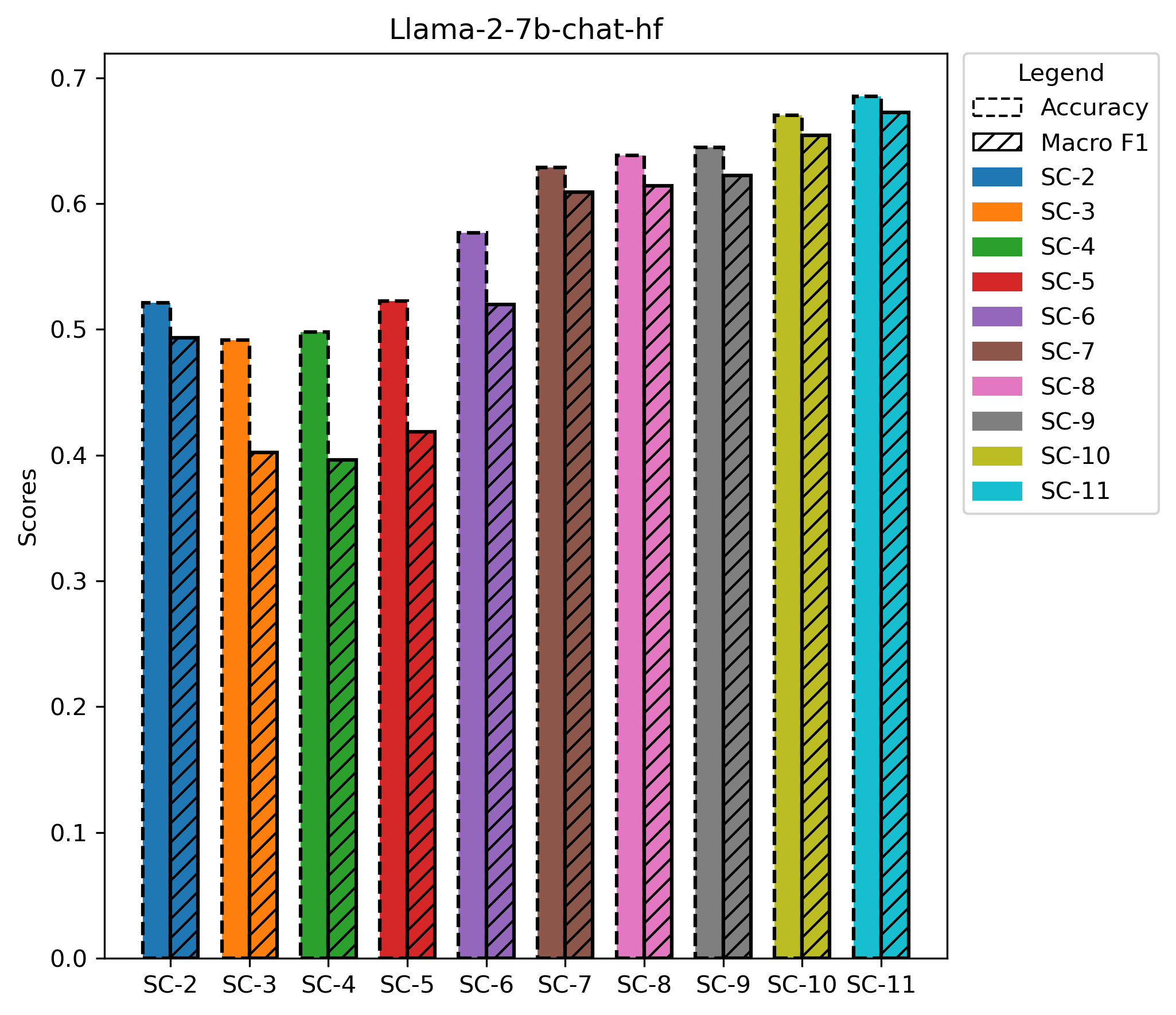}
    \caption{Impact of ensembling context-size-specific models within the \MethodName{} framework on the Subjective dataset in a 16-shot setting. Result is reported for Llama-2-7b-chat-hf, using Accuracy and Macro-F1 scores averaged over 5 random seeds. Each ensemble, denoted SC-\textit{N}, aggregates calibration models trained on context sizes ranging from 1 to $N$ (e.g., SC-2 uses models with context sizes 1 and 2, SC-11 includes context sizes 1 through 11). The consistent improvement in performance as N increases across all three LLMs highlights the general benefit of aggregating insights from a more diverse set of k-shot learners.}
    \label{fig: 16-shot learners Llama}
    
\end{figure}

Second, we investigate whether ensembling calibrators trained with different context sizes improves predictive performance. Concretely, we train a collection of models $\{\LearnedParams^i\}_{i \in I}$, where each calibrator is fitted using training data with $i$ in-context examples. We then ensemble these context-size-specific calibrators and evaluate the impact of increasing the number of distinct $i$-shot learners in the ensemble (i.e., increasing $|I|$). Empirically, we observe a consistent and monotonic improvement in both Accuracy and Macro-F1 scores as $|I|$ grows as shown in Figure \ref{fig: k-shot learners qwen} and \ref{fig: 16-shot learners Llama}. This suggests that calibrators exposed to heterogeneous amounts of contextual information offer complementary signals, enhancing the robustness and predictive accuracy of the final calibrated output. These findings highlight a promising direction: with sufficient computational resources, one could train and ensemble an even broader set of context-specific calibrators to capture a richer diversity of contextual patterns, potentially unlocking further performance gains.

\subsection{Macro-F1 Gains as the number of sampled sub-contexts Increases}\label{eq: sampled sub-contexts}
\begin{figure}[h]
    \centering 
    \includegraphics[width=0.6\columnwidth]{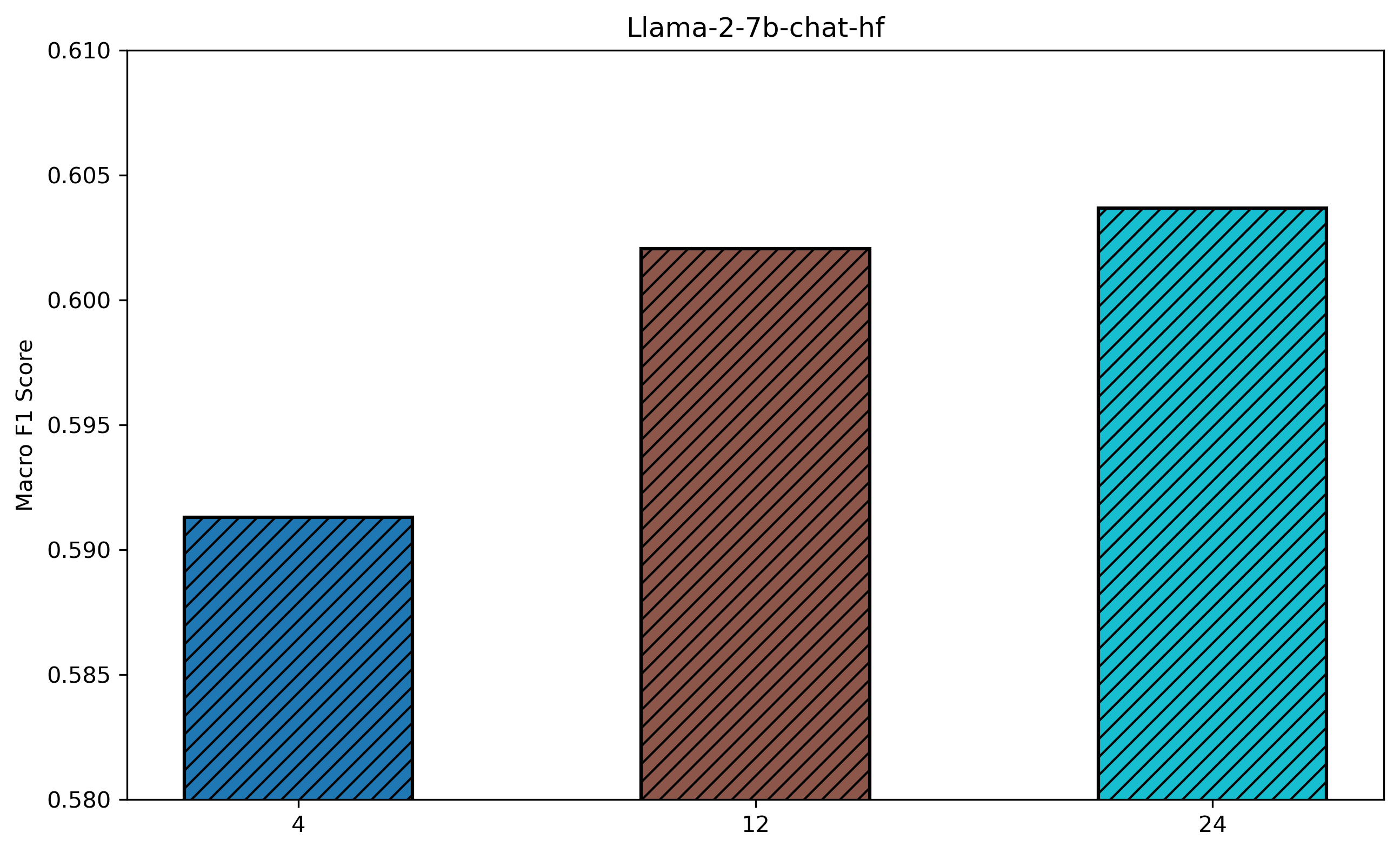}
   \caption{Impact of the number of sampled sub-contexts ($m_i$) used for prediction averaging within each context-size-specific model in the SC ensemble. Results show Macro-F1 scores on the Subjective dataset using the Llama-2-7b-chat-hf model in an 8-shot setting, averaged over 5 random seeds. The x-axis ($m_i$) represents the number of distinct contexts of a given size $i$ sampled to generate predictions, which are then averaged. Performance improves as more context variations are considered in the ensemble prediction.}
    \label{fig: Llama icl inference}
    
\end{figure}

Next, we investigate the impact of the number of sampled sub-contexts ($m_i$) used for prediction averaging within each context-size-specific calibrator during the ensembling phase. In Figure \ref{fig: Llama icl inference}, our findings reveal that increasing $m_i$ (i.e., averaging predictions over a greater number of distinct sub-contexts of size $i$) generally enhances Macro-F1 scores. This suggests that more comprehensive sampling of available context variations for each $i$-shot learner improves the accuracy of the ensemble's output, helping to further reduce ICL's sensitivity to specific context compositions.

\subsection{Compute and timing.}\label{eq: Compute and timing} In Tables \ref{tab:timing_k4} and \ref{tab:timing_k8},
we characterize the computational footprint of sub-context (SC) ensembling by reporting wall-clock training time \(T_{\text{train}}\) and inference time \(T_{\text{infer}}(m_i)\) per 256 test examples, where \(m_i\) is the number of sampled sub-contexts with size $i$ used at inference for \(\mathrm{SC}_{i}\). Training is a one-time cost per method. SC rows are cumulative. Specifically, for \(k=4\) we aggregate \(\mathrm{SC}_{2}\)\textendash{}\(\mathrm{SC}_{3}\), and for \(k=8\) we aggregate \(\mathrm{SC}_{2}\)\textendash{}\(\mathrm{SC}_{5}\), whereas all bias-only baselines are effectively insensitive to \(m_i\).

Specifically, SC ensembling increases inference time approximately linearly with  \(m_i\) because each additional sub-context entails an extra forward pass. This trend is evident at both context sizes. For \(k=4\), combining \(\mathrm{SC}_{2}\) and \(\mathrm{SC}_{3}\) adds a modest \(T_{\text{train}}=2.24\) s and yields \(T_{\text{infer}}(1)=22.91\) s, growing to \(T_{\text{infer}}(6)=134.96\) s, while baselines remain near \(10.5\) s regardless of  \(m_i\). For \(k=8\), the cumulative \(\mathrm{SC}_{2}\)\textendash{}\(\mathrm{SC}_{5}\) configuration requires \(T_{\text{train}}=489.62\) s and exhibits \(T_{\text{infer}}(1)=42.83\) s rising to \(T_{\text{infer}}(6)=260.32\) s, with baselines staying close to \(11.1\) s across all settings. These measurements are consistent with the simple cost model
\[
T_{\text{infer}}(m_i)\;\approx\; m_i\times T_{\text{base},i} \;+\; \text{overhead},
\]
in which \(T_{\text{base}},i\) is the per-example cost of a single forward pass with context size $i$.

Practically speaking, when computation is a limiting factor, running the most effective single SC size offers a favorable accuracy–cost trade-off. In our experiments, \(\mathrm{SC}_{3}\) for \(k=4\) and \(\mathrm{SC}_{5}\) for \(k=8\) are the strongest individual calibrators, preserving most of the ensemble’s accuracy gains while keeping inference overhead substantially closer to baseline runtimes.

\begin{table}[h]
\centering
\caption{Training and inference timing (seconds) for $k=4$.}
\label{tab:timing_k4}
\begin{tabular}{lccccccc}
\toprule
\textbf{Method} & \textbf{T$_{\text{train}}$(s)} & \textbf{T$_{\text{infer}}$(1)} & \textbf{T$_{\text{infer}}$(2)} & \textbf{T$_{\text{infer}}$(3)} & \textbf{T$_{\text{infer}}$(4)} & \textbf{T$_{\text{infer}}$(5)} & \textbf{T$_{\text{infer}}$(6)} \\
\midrule
Baseline & 0.00 & 10.51 & 10.51 & 10.51 & 10.51 & 10.51 & 10.51 \\
CC       & 0.12 & 10.48 & 10.48 & 10.48 & 10.48 & 10.48 & 10.48 \\
Domain   & 0.85 & 10.47 & 10.47 & 10.47 & 10.47 & 10.47 & 10.47 \\
Batch    & 0.00 & 10.54 & 10.54 & 10.54 & 10.54 & 10.54 & 10.54 \\
\midrule
SC$_2$   & 1.26 & 12.52 & 24.44 & 36.65 & 49.66 & 61.09 & 73.15 \\
SC$_3$   & 0.98 & 10.39 & 20.85 & 31.15 & 42.74 & 52.14 & 61.81 \\
\textbf{SC} & \textbf{2.24} & \textbf{22.91} & \textbf{45.29} & \textbf{67.80} & \textbf{92.40} & \textbf{113.23} & \textbf{134.96} \\
\bottomrule
\end{tabular}
\end{table}

\begin{table}[h]
\centering
\caption{Training and inference timing (seconds) for $k=8$.}
\label{tab:timing_k8}
\begin{tabular}{lccccccc}
\toprule
\textbf{Method} & \textbf{T$_{\text{train}}$(s)} & \textbf{T$_{\text{infer}}$(1)} & \textbf{T$_{\text{infer}}$(2)} & \textbf{T$_{\text{infer}}$(3)} & \textbf{T$_{\text{infer}}$(4)} & \textbf{T$_{\text{infer}}$(5)} & \textbf{T$_{\text{infer}}$(6)} \\
\midrule
Baseline & 0.00 & 11.34 & 11.34 & 11.34 & 11.34 & 11.34 & 11.34 \\
CC       & 0.13 & 11.11 & 11.11 & 11.11 & 11.11 & 11.11 & 11.11 \\
Domain   & 0.95 & 11.14 & 11.14 & 11.14 & 11.14 & 11.14 & 11.14 \\
Batch    & 0.00 & 11.13 & 11.13 & 11.13 & 11.13 & 11.13 & 11.13 \\
\midrule
SC$_2$   & 16.08 & 11.75 & 24.15 & 36.01 & 47.80 & 59.14 & 71.58 \\
SC$_3$   & 66.44 & 10.11 & 21.03 & 31.60 & 41.92 & 52.19 & 63.09 \\
SC$_4$   & 201.79 & 10.37 & 20.69 & 31.13 & 41.65 & 52.05 & 61.95 \\
SC$_5$   & 205.31 & 10.60 & 21.02 & 32.00 & 42.17 & 52.96 & 63.70 \\
\textbf{SC} & \textbf{489.62} & \textbf{42.83} & \textbf{86.89} & \textbf{130.74} & \textbf{173.54} & \textbf{216.34} & \textbf{260.32} \\
\bottomrule
\end{tabular}
\end{table}

\subsection{Effects of Trust-Region and Invariance}\label{app: Trust-Region and Invariance}
To isolate the impact of the key components of our proposed method, we conduct an ablation study, with the results presented in Table \ref{tab:ablation}. We evaluate the performance contributions of our two main components: the directional trust-region constraint and the context invariance penalty.

The study begins with the "Uncalibrated (Baseline)" model, which achieves a Macro-F1 of 0.634. Introducing the core calibration mechanism without our proposed constraints ("No trust-region, no invariance") already yields a substantial improvement. When adding either the "Invariance only" or "Trust-region only" component, performance increases further, with both contributing similarly to the overall score. However, the full model, which combines both trust-region + invariance, achieves the highest performance across both Macro-F1 (0.746) and Accuracy (0.788). This demonstrates that both components are crucial and complementary, working together to deliver the best calibration results.

\begin{table}[h]
\centering
\caption{Ablation study on the components of \MethodName{}. Results show Macro-F1 and Accuracy, reported as mean $\pm$ standard error.}
\label{tab:ablation}
\begin{tabular}{lcc}
\toprule
\textbf{Method} & \textbf{Macro-F1 $\pm$ SE} & \textbf{Accuracy $\pm$ SE} \\
\midrule
Uncalibrated (Baseline) & $0.634 \pm 0.008$ & $0.759 \pm 0.008$ \\
No trust-region, no invariance & $0.695 \pm 0.056$ & $0.729 \pm 0.047$ \\
Invariance only & $0.705 \pm 0.063$ & $0.741 \pm 0.054$ \\
Trust-region only & $0.706 \pm 0.060$ & $0.743 \pm 0.049$ \\
\textbf{Both: trust-region + invariance} & \textbf{0.746 $\pm$ 0.041} & \textbf{0.788 $\pm$ 0.030} \\
\bottomrule
\end{tabular}
\end{table}

\subsection{Scaling to Larger Models (LLaMA-13B)}\label{app: Larger Models}

To assess the scalability of our method, we ran additional experiments with the larger LLaMA-13B model. Due to computational constraints, we focused this scaling analysis on three datasets, \textbf{Rotten Tomatoes}, \textbf{SST-2}, and \textbf{AGNews}, where we compared its performance against the 7B variant. All experiments were conducted under the same 4-shot setup and averaged over 5 random seeds.

The results, presented in Tables \ref{tab:rotten_tomatoes_13b}, \ref{tab:sst2_13b}, and \ref{tab:agnews_13b}, demonstrate that our method, SC, scales effectively. Across all three datasets, SC consistently delivers the strongest performance on the LLaMA-13B model, achieving the highest Macro-F1 and Accuracy. Notably on AGNews, while the 7B baseline was competitive, SC provides a substantial improvement for the 13B model, boosting accuracy from 78.12 to 88.05. This confirms that our calibration approach remains highly effective and provides consistent benefits as the underlying language model size increases. We plan to incorporate further evaluations on even larger models in future work.

\begin{table}[h!]
\centering
\caption{Performance on the Rotten Tomatoes dataset with 7B and 13B models.}
\label{tab:rotten_tomatoes_13b}
\begin{tabular}{lcccc}
\toprule
\textbf{Method} & \textbf{Macro-F1 (7B) $\pm$ SE} & \textbf{Accuracy (7B) $\pm$ SE} & \textbf{Macro-F1 (13B) $\pm$ SE} & \textbf{Accuracy (13B) $\pm$ SE} \\
\midrule
Baseline & $91.16 \pm 1.59$ & $91.17 \pm 1.58$ & $91.87 \pm 0.48$ & $91.89 \pm 0.49$ \\
CC       & \textbf{92.06 $\pm$ 0.96} & \textbf{92.07 $\pm$ 0.95} & $92.33 \pm 0.11$ & $92.38 \pm 0.11$ \\
DC       & $91.92 \pm 1.13$ & $91.95 \pm 1.18$ & $92.25 \pm 0.12$ & $92.29 \pm 0.10$ \\
Batch    & $91.52 \pm 1.25$ & $91.52 \pm 1.28$ & $91.38 \pm 0.59$ & $91.41 \pm 0.57$ \\
SC       & $91.56 \pm 1.19$ & $91.57 \pm 1.18$ & \textbf{92.33 $\pm$ 0.26} & \textbf{92.38 $\pm$ 0.25} \\
\bottomrule
\end{tabular}
\end{table}

\begin{table}[h!]
\centering
\caption{Performance on the SST-2 dataset with 7B and 13B models.}
\label{tab:sst2_13b}
\begin{tabular}{lcccc}
\toprule
\textbf{Method} & \textbf{Macro-F1 (7B) $\pm$ SE} & \textbf{Accuracy (7B) $\pm$ SE} & \textbf{Macro-F1 (13B) $\pm$ SE} & \textbf{Accuracy (13B) $\pm$ SE} \\
\midrule
Baseline & $93.36 \pm 0.44$ & $93.63 \pm 0.40$ & $95.10 \pm 0.56$ & $95.21 \pm 0.56$ \\
CC       & $93.17 \pm 1.92$ & $93.49 \pm 0.91$ & $94.81 \pm 0.74$ & $94.92 \pm 0.73$ \\
DC       & $93.60 \pm 0.50$ & $93.83 \pm 1.19$ & $95.47 \pm 0.09$ & $95.61 \pm 0.10$ \\
Batch    & $93.53 \pm 0.47$ & $93.75 \pm 0.48$ & $95.42 \pm 0.65$ & $95.51 \pm 0.65$ \\
SC       & \textbf{94.25 $\pm$ 0.53} & \textbf{94.45 $\pm$ 0.47} & \textbf{95.65 $\pm$ 0.26} & \textbf{95.80 $\pm$ 0.25} \\
\bottomrule
\end{tabular}
\end{table}

\begin{table}[h!]
\centering
\caption{Performance on the AGNews dataset with 7B and 13B models.}
\label{tab:agnews_13b}
\begin{tabular}{lcccc}
\toprule
\textbf{Method} & \textbf{Macro-F1 (7B) $\pm$ SE} & \textbf{Accuracy (7B) $\pm$ SE} & \textbf{Macro-F1 (13B) $\pm$ SE} & \textbf{Accuracy (13B) $\pm$ SE} \\
\midrule
Baseline & \textbf{77.58 $\pm$ 7.17} & \textbf{82.58 $\pm$ 4.17} & $85.74 \pm 1.77$ & $87.19 \pm 1.27$ \\
CC       & $71.01 \pm 3.42$ & $79.30 \pm 2.02$ & $66.40 \pm 0.61$ & $77.73 \pm 0.28$ \\
DC       & $72.10 \pm 3.61$ & $79.61 \pm 1.89$ & $66.90 \pm 1.00$ & $77.81 \pm 0.60$ \\
Batch    & $66.06 \pm 2.94$ & $77.19 \pm 1.27$ & $66.32 \pm 0.63$ & $77.58 \pm 0.29$ \\
SC       & $71.76 \pm 11.31$ & $78.12 \pm 8.67$ & \textbf{87.51 $\pm$ 1.13} & \textbf{88.05 $\pm$ 0.94} \\
\bottomrule
\end{tabular}
\end{table}

\end{document}